\documentclass[10pt,twocolumn,letterpaper]{article}

\usepackage{cvpr}              % To produce the CAMERA-READY version
\usepackage{graphicx}
\usepackage{amsmath}
\usepackage{amssymb}
\usepackage{booktabs}
\usepackage[accsupp]{axessibility}

\usepackage[pagebackref,breaklinks,colorlinks]{hyperref}

\usepackage[capitalize]{cleveref}
\crefname{section}{Sec.}{Secs.}
\Crefname{section}{Section}{Sections}
\Crefname{table}{Table}{Tables}
\crefname{table}{Tab.}{Tabs.}

\def\cvprPaperID{23} % *** Enter the CVPR Paper ID here
\def\confName{CVPR}
\def\confYear{2022}
\begin{document}

%%%%%%%%% TITLE - PLEASE UPDATE
\title{Color Invariant Skin Segmentation}
%\title{\LaTeX\ Author Guidelines for \confName~Proceedings}

% \author{Han Xu $^{1}$, Abhijit Sarkar $^{2}$, Lynn Abbott $^{1}$\\
% $^{1}$Bradley Department of Electrical and Computer Engineering, $^{2}$Virginia Tech Transportation Institute \\ Virginia Tech, Blacksburg, Virginia, USA\\
% {\tt\small \{xuhan1996,asarkar1,abbott\}@vt.edu}

% \author{Han Xu $^{1}$, Abhijit Sarkar $^{2}$, Lynn Abbott $^{1}$\\
% Bradley Department of Electrical and Computer Engineering\\
% Virginia Polytechnic Institute and State University\\
% Blacksburg, VA, USA,
% {\tt\small \{xuhan1996, abbott, asarkar1\}@vt.edu}

\author{Han Xu$^{1}$, Abhijit Sarkar$^{2}$, A. Lynn Abbott$^{1}$\\
$^{1}$ Bradley Department of Electrical and Computer Engineering, \\
$^{2}$ Virginia Tech Transportation Institute\\
Virginia Tech, Blacksburg, VA, USA\\
{\tt\small \{xuhan1996, asarkar1, abbott\}@vt.edu}
% For a paper whose authors are all at the same institution,
% omit the following lines up until the closing ``}''.
% Additional authors and addresses can be added with ``\and'',
% just like the second author.
% To save space, use either the email address or home page, not both
% \and
% A. Lynn Abbott\\
% Bradley Department of Electrical and Computer Engineering, Virginia Polytechnic Institute and State University\\
% Blacksburg, VA, USA\\
% % First line of institution2 address\\
% {\tt\small secondauthor@i2.org}
% \and
% Abhijit Sarkar\\
}
\maketitle

% %%%%%%%%% ABSTRACT
% \begin{abstract}
% This paper addresses the problem of automatically detecting human skin in images without reliance on color information. 
% Although most previous skin-detection methods have used color cues almost exclusively, we present a new approach that performs well in the absence of such information. 
% A key aspect of the work is dataset repair through augmentation that is applied strategically during training, with the goal of color invariant feature learning to enhance generalization.
% We have demonstrated the concept using two architectures, and experimental results show improvements in both precision and recall for most Fitzpatrick skin tones in the ECU dataset.
% We further tested the system with the RFW dataset to show that the proposed method performs much more consistently across different ethnicities, thereby reducing the chance of bias based on skin color.
% Extensive experiments on grayscale images and images under unconstrained illuminations and artificial filters were conducted to demonstrate the effectiveness of our work.
% Source code will be provided with the final version of this paper.
% \end{abstract}

%%%%%%%%% BODY TEXT
\begin{abstract}
This paper addresses the problem of automatically detecting human skin in images without reliance on color information.
A primary motivation of the work has been to achieve results that are consistent across the full range of skin tones, even while using a training dataset that is significantly biased toward lighter skin tones.
Previous skin-detection methods have used color cues almost exclusively, and we present a new approach that performs well in the absence of such information. 
A key aspect of the work is dataset repair through augmentation that is applied strategically during training, with the goal of color invariant feature learning to enhance generalization.
We have demonstrated the concept using two architectures, and experimental results show improvements in both precision and recall for most Fitzpatrick skin tones in the benchmark ECU dataset.
We further tested the system with the RFW dataset to show that the proposed method performs much more consistently across different ethnicities, thereby reducing the chance of bias based on skin color.
To demonstrate the effectiveness of our work, extensive experiments were performed on grayscale images as well as images  obtained under unconstrained illumination and with artificial filters.
Source code: https://github.com/HanXuMartin/Color-Invariant-Skin-Segmentation
\end{abstract}
%%{-.5cm}
%%%%%%%%% BODY TEXT

\section{Introduction}
\begin{figure}
\begin{center}
\includegraphics[width=1\linewidth]{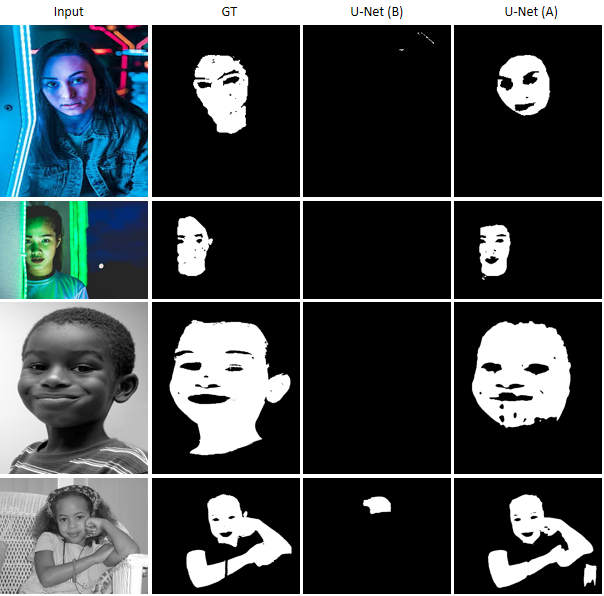}
\end{center}
   \vspace{-.5cm}
   \caption{Example results from our skin-detection system, including cases of complex illumination and imaging conditions. \textit{Left to right:} Input image; ground truth; 
   %U-Net output before training with our augmentation approach; U-Net output after training using our novel augmentation approach. 
   segmentation results using baseline system;
   results after training using our novel augmentation approach. Dramatic improvements in detected skin regions are apparent for all of these cases.
   }
\vspace{-.5cm}
\label{fig:SkinApparanceVariation}
\end{figure}
Skin detection refers to the process of identifying pixels that correspond to human skin within image data or video data.
Automated skin detection can play an important role for sensitive applications such as face detection and recognition (e.g.,~\cite{kovac2003human,liu2010robust}), facial expression recognition, gesture recognition~\cite{alsheakhali2011hand}, content-based image retrieval,  filtering of objectionable content~\cite{forsyth1996finding,drimbarean2001image}, skin rendering in computer graphics~\cite{borshukov2005realistic, donner2008layered}, and virtual reality. 
Although the last two decades have seen many efforts related to  skin detection and skin modeling~\cite{igarashi2005appearance, zollhofer2018state}, it is interesting that almost all  techniques for image-based skin detection depend heavily on the use of color information. Extensive surveys are provided by Mahmoodi \etal~\cite{mahmoodi2016comprehensive} and Kakumanu \etal~\cite{kakumanu2007survey}. 
\par
We contend that over-reliance on color cues has imposed performance limits and has also led to bias related to skin tones.
A major reason for such bias is dataset imbalance, with the large majority of training samples representing lighter skin tones. Additionally,  imaging methods may also introduce variability, including spectral range of sensor arrays (grayscale, near-infrared, RGB) and  creative filtering in photography applications (e.g., sepia tones in movies, or Instagram filters). Examples of such variations  are shown in Figure~\ref{fig:SkinApparanceVariation}. \par
To address such problems, this paper introduces a new technique for human skin detection that significantly reduces reliance on  color information and focuses much  more on texture and contextual information to detect the skin pixels in an image. A significant aspect of the system is that color-space augmentation is applied strategically to the training set so that a resulting deep neural network suppresses the system's dependence on color cues.
Hence, our high-level strategy has been to guide the training procedure away from color cues and toward features related to visual texture and context.
Figure \ref{fig:Schematic} provides an illustration of the augmentation strategy and the training strategy.
We demonstrate our procedure by training the U-Net architecture~\cite{ronneberger2015u} and FCN~\cite{long2015fully} using ECU \cite{phung2005skin} dataset and do testing on both ECU and RFW \cite{wang2019racial} dataset. 
% The ECU dataset is a common benchmark dataset  for skin detection. 
% The RFW dataset contains face images with annotations representing four ethnic groups. We have used the RFW dataset to show that our new algorithm shows virtually no bias to any ethnicity and race. 

The primary contributions of this paper are as follows. 
\textit{Color invariance.} We describe an approach to automated detection of human skin that does not depend on the color properties of the skin. %Further, our approach performs well even in the presence of significant dataset imbalance.
%does not require additional costly datasets or annotations.  
\textit{Universality.} The resulting system therefore has  potential to operate in environments with relatively unconstrained illumination conditions, including extreme cases of over- and underexposed images, grayscale images, and systems that utilize  with creative filters (such as Instagram). As such, the system is intended for operation ``in the wild,'' and can  relax requirements and reduce costs related to camera selection.
\textit{Little or no racial bias.} In our experimental results, we have systematically evaluated the performance of our algorithm   for subjects with different skin tones. Using cross-database testing, we have shown that our new algorithm performs virtually uniformly across all of the available annotated skin tones.\par
Data imbalance is a typical problem in data driven models. Hence, we need to understand the bias before hand and use intelligent algorithm. It is our hope that our color-augmentation strategy for training and testing can be applied widely to other domains, in order to address problems related to racial and social bias. 
%-------------------------------------------------------------------------
\section{Related work}

\subsection{Skin detection for natural images }

Early approaches to skin detection focused primarily on the use of color cues (e.g., \cite{vezhnevets2003survey,phung2005skin,kakumanu2007survey}), with the goal of detecting different skin tones under varying illumination conditions.  
In a few cases, researchers also incorporated cues related to visual texture (e.g., \cite{garcia1999face,cula2005skin,fotouhi2009skin}) or shape (e.g., \cite{drimbarean2001image})  as a supplement to color information.
In one case, researchers considered texture and contextual cues without the use of color~\cite{sarkar2017universal}.
More recently, researchers have applied methods using deep neural networks to the problem of skin segmentation. 
The different approaches may be grouped loosely into 3 categories: FCN-based ~\cite{long2015fully}, R-CNN-based ~\cite{girshick2014rich,he2017mask}, and encoder-decoder models \cite{ronneberger2015u,badrinarayanan2017segnet}. 

Under the first category, 
Zuo \etal~\cite{zuo2017combining} introduced an end-to-end network for human skin detection by integrating an RNN (Recurrent Neural Network) into an FCN model. They were able to demonstrate improved skin-detection performance in complex environments, including ECU and COMPAQ dataset \cite{jones2002statistical}.
He \etal~\cite{he2019semi} later proposed a  semisupervised skin-detection method to address the problem of insufficient training samples. Compared to some state-of-art methods.
Within the second category, 
Roy \etal~\cite{roy2017deep} used an R-CNN-based approach to reduce the number of false positives by adding a CNN-based skin detector. This approach yielded a substantial improvement over a baseline of using R-CNN only.
Nguyen \etal~\cite{nguyen2018hand} integrated a mean-shift hand tracker into Mask R-CNN~\cite{he2017mask}.
They reported improvements of  5$\%$ to 9$\%$ in detection accuracy, compared to Mask R-CNN alone.
Under the encoder-decoder category,
Nguyen \etal~\cite{nguyen2019hand} modified the original SegNet \cite{badrinarayanan2017segnet} architecture by increasing the number of decoders, thereby allowing each encoder to perform multiple tasks at the same time, which discriminate skin components in the hand area more accurately. 
Topiwala~\cite{8941944} has shown that U-Net stands out among the frequently-used skin detectors on their  dataset of the human abdomen with different skin colors,
The method based on U-Net was also computationally faster.
Tarasiewicz~\cite{tarasiewicz2020skinny} refined the U-Net architecture~\cite{ronneberger2015u}
by considering large-scale contextual features, using inception and dense blocks to reduce occurrences of false positives significantly while doing skin detection. 

% \vspace{-.5cm}
\subsection{Algorithmic bias}

This work has been motivated in part by the need to promote demographic fairness in automated systems, particularly relating to differences in skin tones that are related to ethnicity and race. %ethnicity, race, gender, age, and other respects. 
For tasks such as face recognition,  techniques have been developed recently to evaluate bias within algorithms and datasets~\cite{buolamwini2018gender},
and to improve fairness with respect to such differences~\cite{drozdowski2020demographic}.
More generally, Mehrabi \etal~\cite{mehrabi2021survey} have surveyed the Machine Learning field and have
developed a taxonomy of problems that affect bias and fairness within automated systems. Most bias-mitigation systems focus on two types of biases: dataset bias, and task bias. This paper is concerned with the former, which refers to  datasets having classes  are not represented as well as others within the dataset. 
Researchers recently have focused on invariant feature learning for protected variables (here, the skin color appearance), and perform database repair to eliminate the representation error~\cite{adeli2021representation,li2019repair,salimi2019interventional}. 
This paper uses the database repair approach through augmentation for de-biasing. 

%-------------------------------------------------------------------------

\begin{figure*}
\centering
\begin{subfigure}{0.38\textwidth}
\begin{center}
\includegraphics[width=1.0\linewidth]{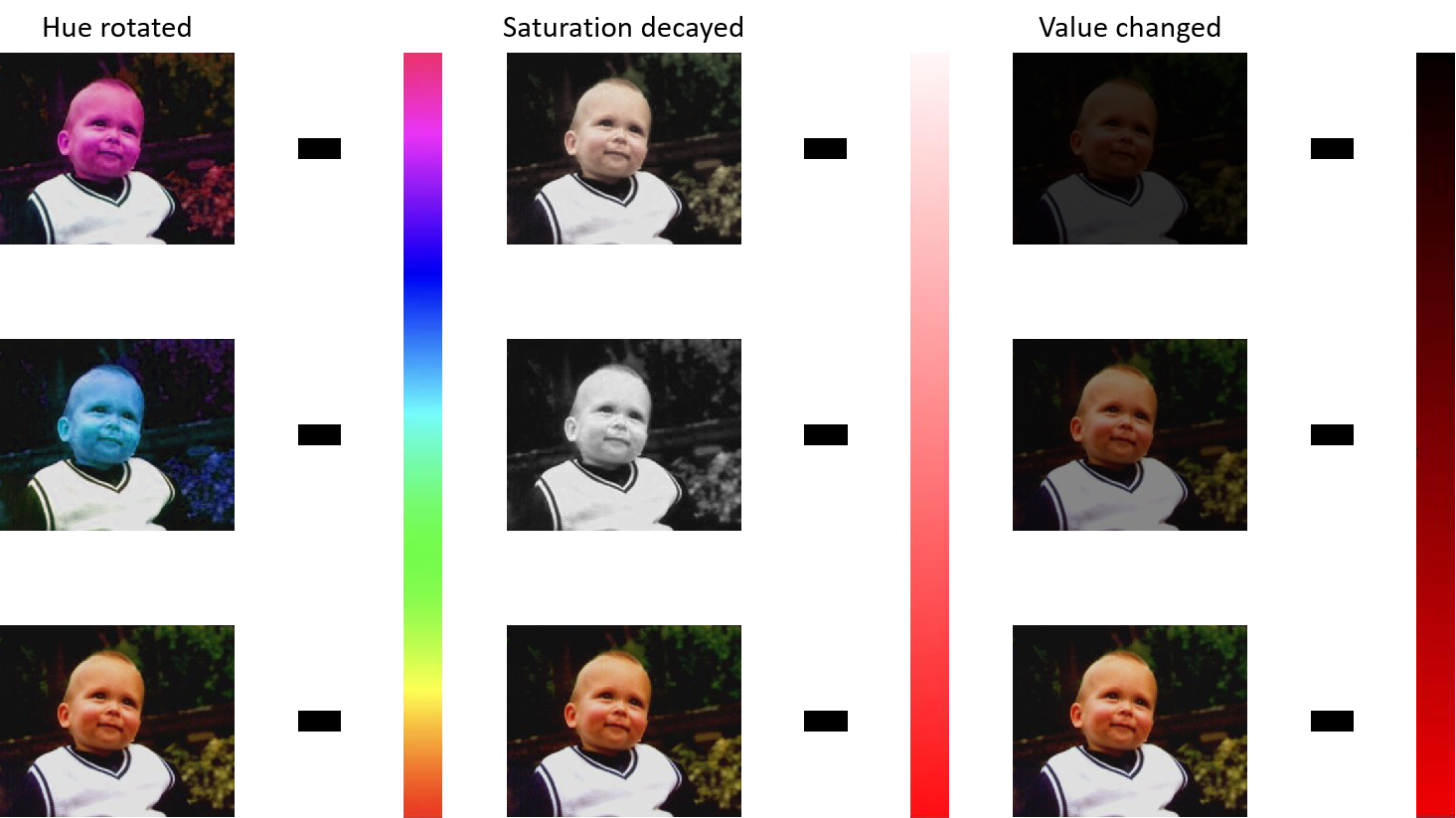}
\end{center}
\caption{}
\label{fig:color space augmentation}
\end{subfigure}
\begin{subfigure}{0.6\textwidth}
\includegraphics[width=1.0\linewidth]{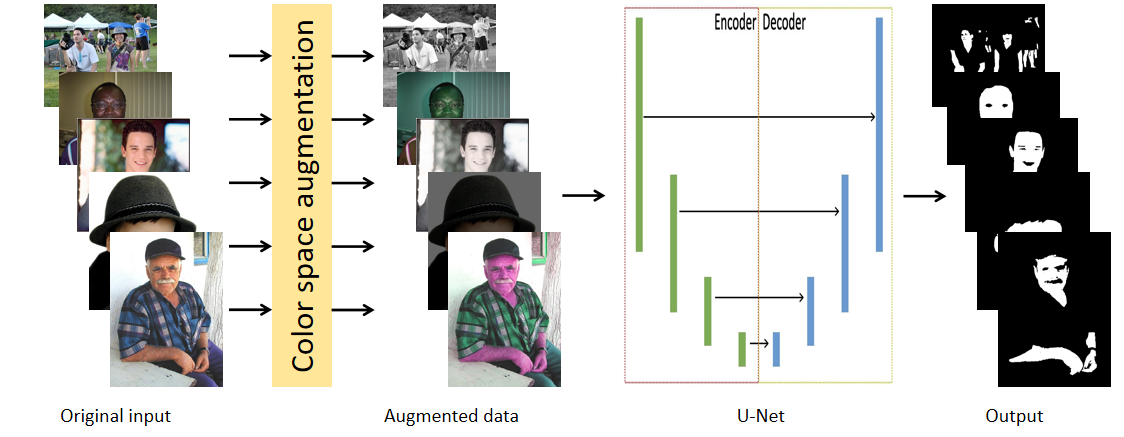}
\caption{}
\label{fig:overall structure}
\end{subfigure}
\vspace{-0.35cm}
\caption{
Schematic diagram showing the overall approach. (a) Color space augmentation in HSV  space containing hue rotation, saturation decay and value change. (b) Overall structure of our skin detection system. During training, each input image undergoes color augmentation.
}
\vspace{-0.5cm}
\label{fig:Schematic}
\end{figure*}
\section{Methods}

The overall process is described in Figure \ref{fig:Schematic}. We first use an image augmentation approach to create an expanded dataset (Figure \ref{fig:color space augmentation}), and use the dataset for training a U-Net-based segmentation network for skin detection (Figure \ref{fig:overall structure}). 
\begin{figure}
\centering
\begin{subfigure}{0.23\textwidth}
\centering
\includegraphics[width=1\linewidth]{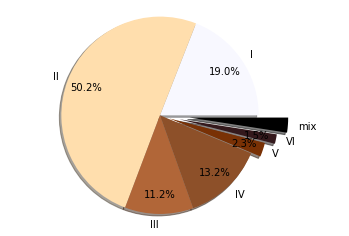} 
\caption{}
\label{fig:ECU skin tones distribution}
\end{subfigure}
\begin{subfigure}{0.23\textwidth}
\centering
\includegraphics[width=0.5\linewidth]{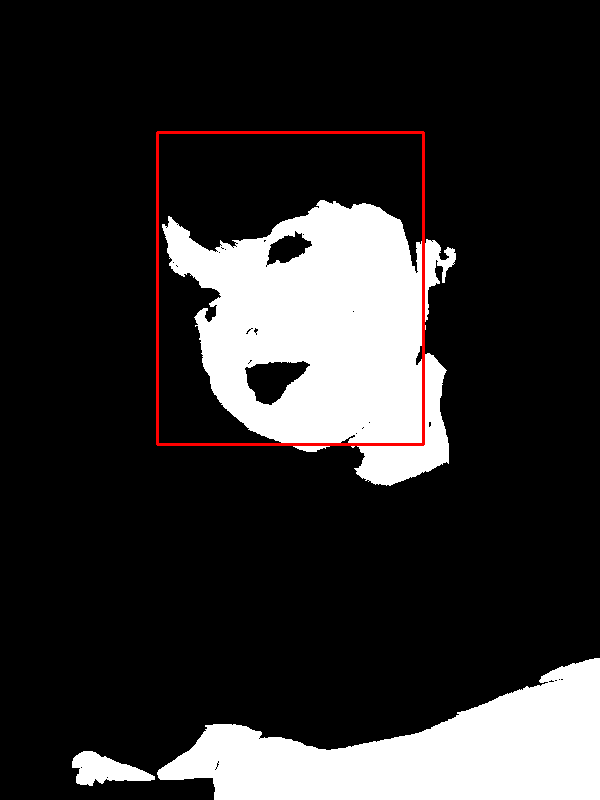}
\caption{}
\label{fig:skin in face gt}
\end{subfigure}
\vspace{-0.35cm}
\caption{(a) Distribution of skin types in the ECU dataset. Labels I-VI refer to the six skin tones described by Fitzpatrick~\cite{fitzpatrick1988validity}. The group ``mix'' refers to several skin tone categories in a single image. %This uneven distribution may result in bias.
(b) An example of skin/face evaluation, as described in equation (\ref{fun: skin /face}). 
}
\vspace{-.5cm}
\label{fig:skin /face}
\end{figure}
\subsection{Dataset repair by augmentation}
% Data augmentation has been widely used in computer vision, especially for the purpose of regularizing the network \cite{wang2021regularizing, mikolajczyk2018data}. However, the augmentation techniques are also useful for reducing generalization error \cite{tellez2019quantifying,mikolajczyk2018data} through dataset repair \cite{li2019repair}. These include noise augmentation \cite{ding2016convolutional}, color augmentation\cite{tellez2019quantifying}, pose augmentation etc. 
In this work, we adopt color-based data augmentations that can add artificial images to mimic alternate representations of the image, in this case, the appearance of the skin. 
We have implemented augmentation in the HSV (Hue, Saturation, Value) space. 
In general, the ambient illumination, especially the amount of light reflected by the skin surface, is reflected in the value channel. It includes shadows and exposure. Saturation indicates  the purity of the color. On the other hand, the spectral property of the illumination source is reflected in the hue channel and the saturation channel. (Examples are shown in Figure \ref{fig:SkinApparanceVariation}.)

\begin{figure*}
\begin{subfigure}{1\textwidth}
\includegraphics[width=1\linewidth]{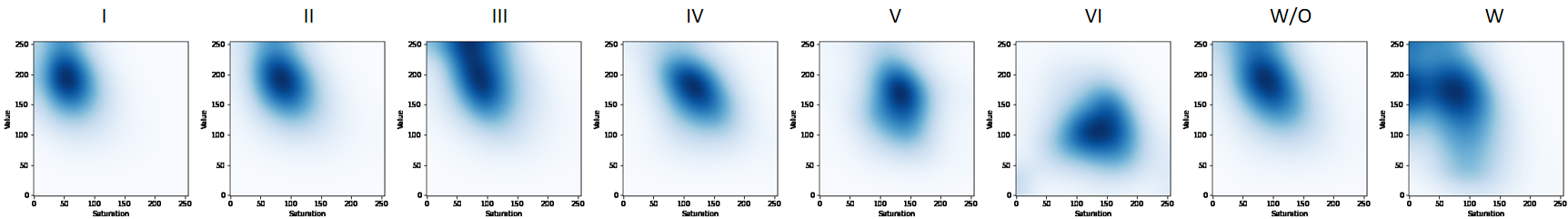}
\caption{}
\end{subfigure}
\begin{subfigure}{1\textwidth}
\includegraphics[width=1\linewidth]{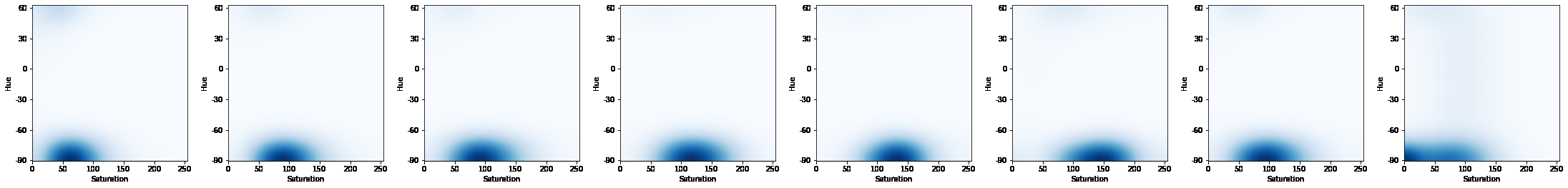}
\caption{}
\end{subfigure}
\begin{subfigure}{1\textwidth}
\includegraphics[width=1\linewidth]{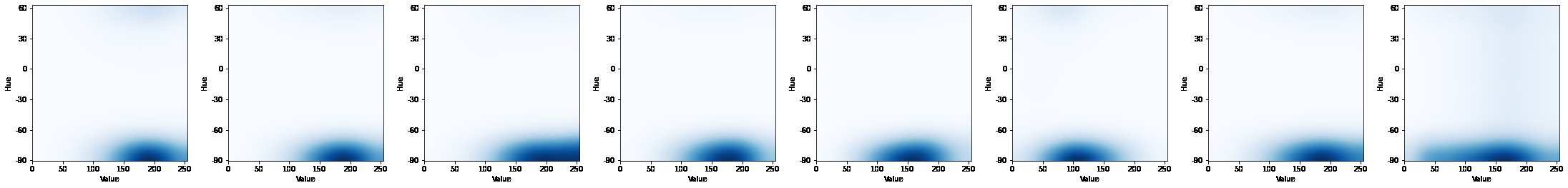}
\caption{}
\end{subfigure}
\vspace{-0.35cm}
\caption{Heatmaps from ECU dataset in three dimensions: (a) Saturation-Value, (b) Saturation-Hue, (c) Value-Hue. The first six columns mark the skin pixel distributions of Fitzpatrick \cite{fitzpatrick1988validity} skin tones I-VI. The last two columns refer to the skin pixel distribution of the training set before (W/O) and after (W) our color space augmentation.}
\vspace{-0.35cm}
\label{fig:heatmap}
\end{figure*}

%Besides, 
Studies  have shown that  physiological biases, particularly skin tones, can influence computer the development of vision algorithms~\cite{sixta2020fairface,buolamwini2018gender,buolamwini2017gender}.
To illustrate a potential cause of  bias, Figure \ref{fig:heatmap} provides heatmaps from  the ECU testing set. 
%These plots show the distributions 
%We illustrate the heatmap from the testing set of the ECU dataset in Figure \ref{fig:heatmap} to interpret how skin tones % \cite{fitzpatrick1988validity} 
%affect the appearance of skin pixels. 
We first classified the those images %testing set in the ECU dataset 
into six skin-tone categories according to~\cite{fitzpatrick1988validity}. 
Then every image in the testing set was converted into HSV color space, and its skin pixels were allocated into different bins according to the (S, V), (S, H), and (V, H) value pairs. The first six columns show the distributions of skin pixels for the different Fitzpatrick categories %of different skin tones 
in HSV color space. 
Column 7 in the figure (``W/O'') shows the composite distribution for the dataset, and it is clearly seen that %HSV values for 
the darkest skin tones are poorly represented within HSV space. 
%Often due to less representation from darker skin, a dataset fails to cover all six skin areas (column 7, ``W/O''). 
Our recommended method extends that representation and aims towards a  distribution that is much more balanced across all classes. All three rows in column 8 (``W'') show improvement. 

Figure \ref{fig:HSV augmentation and result} illustrates how color augmentation works in our experiments.  We augmented the training set with three groups of thresholds in H, S, and V channels respectively. Each image in the training set will be converted into fifteen images, so the original training set will be fifteen times larger. This augmented training set will be used for training the skin segmentation models. The thresholds we selected in the experiments are listed in the figure \ref{fig:HSV augmentation and result}.
%%%%%%%%%%%%%%%%%%%%%%%%%%%%%%%%%%%%%%%%%%%%%%%%%%%%%%%%%%%%%%%%%%%%%%%%%%%%%%%%%%%%%

\subsection{Training segmentation networks}

% The input to the U-Net is an RGB image, and the output is a binary mask (skin = 1, non-skin = 0). 
The  experiment uses NVIDIA GeForce RTX 2070 SUPER GPU with 16 GB GPU memory. The algorithm is trained and tested with U-Net\cite{opensourseUNet} and FCN\cite{opensourseFCN}.The U-Net is working under Python 3.8.5 and Tensorflow 2.3.1 environment with no pre-trained models. 
The FCN is working under Python 3.8.5 and Pytorch 1.70 with a pre-trained VGG-16 network.
% The networks can take input images of size 256$\times$256 pixels, producing the same size's detection results. 
We train the network with a fixed learning rate of 10$^{-4}$, and each epoch takes around 79s when batch size is set to 8. The network will use a module ImageDataGenerator\cite{keras} in the keras to do data augmentation, which includes image rotation, width shift, height shift, shear, zoom, and horizontal flip with nearest fill mode. We used binary cross-entropy loss:
\begin{equation}
L = -\frac{1}{N}\sum_{i=1}^{N}y_{i}\log(f(y_{i}))+(1-y_{i})\log(1-f(y_{i}))
\end{equation}
where $N$ is the number of segmentation classes. 
The symbol $y_{i}$ is the label and $f(y_{i})$ is the predicted probability of the points belonging to the $i^{th}$ class.
The original output of the network will be from 0 to 1. Since the pixels should belong to either skin category or non-skin category, we use the function below to make the output $\mathcal{O}$ binary where 1 refers to skin pixels. The threshold $\delta$ we set is 0.5. 
To make the experiments more convincing, we also draw precision-recall curve in the supplementary materials. 
\begin{equation}
\mathcal{O}=\left\{
\begin{aligned}
&1 && {\mathrm{if}\ f(y_{i})\geq \delta}\\
&0 && {\mathrm{if}\ f(y_{i})<\delta}
\end{aligned}
\right.
\end{equation}
% %%{-.5cm}
\subsection{Datasets}
\begin{figure*}
\centering
\begin{subfigure}{0.31\textwidth}
\includegraphics[width=1.0\linewidth]{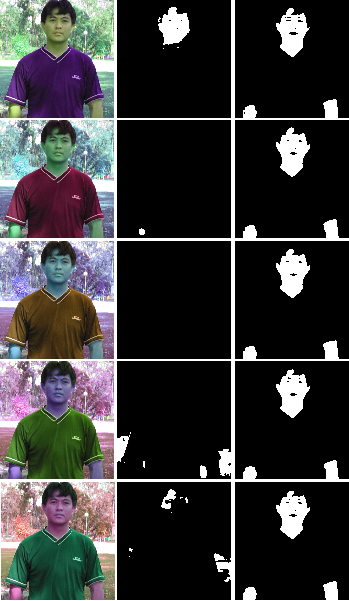} 
\caption{}
\end{subfigure}
\quad
\begin{subfigure}{0.31\textwidth}
\includegraphics[width=1.0\linewidth]{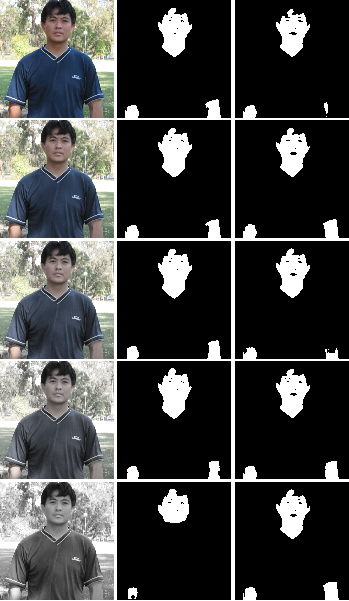}
\caption{}
\end{subfigure}
\quad
\begin{subfigure}{0.31\textwidth}
\includegraphics[width=1.0\linewidth]{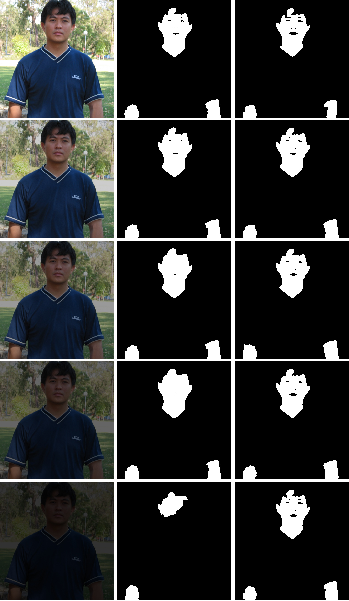}
\caption{}
\end{subfigure}
\vspace{-0.35cm}
\caption{Example of color augmentation across \textit{hue} (a), \textit{saturation} (b), and \textit{value} (c). The first column of each group shows the changed images $\mathcal{I}_{new}$. The second columns show the skin segmentation results without color space augmentation. The third columns show the results with color space augmentation. The input images are rotated at every 60 degrees in the hue channel in the group (a). For group (b), the saturation of images are decayed at  ratios of (0.8, 0.6, 0.4, 0.2, 0.0). For group (c), the values of the images are changed at  ratios of (1.0, 0.8, 0.6, 0.4, 0.2).}
\label{fig:HSV augmentation and result}
\vspace{-0.35cm}
\end{figure*}
%%%%%%%%%%%%%%%%%%%%%%%%%%%%%%%%%%%%%%%%%%%%%%%%%%%%%%%%%%%%%%%%%%%%%%%%%%%%%%%%%%%%%%%%%%%%%%%%%%%
\begin{figure*}[t]
\centering
\includegraphics[width=1\linewidth]{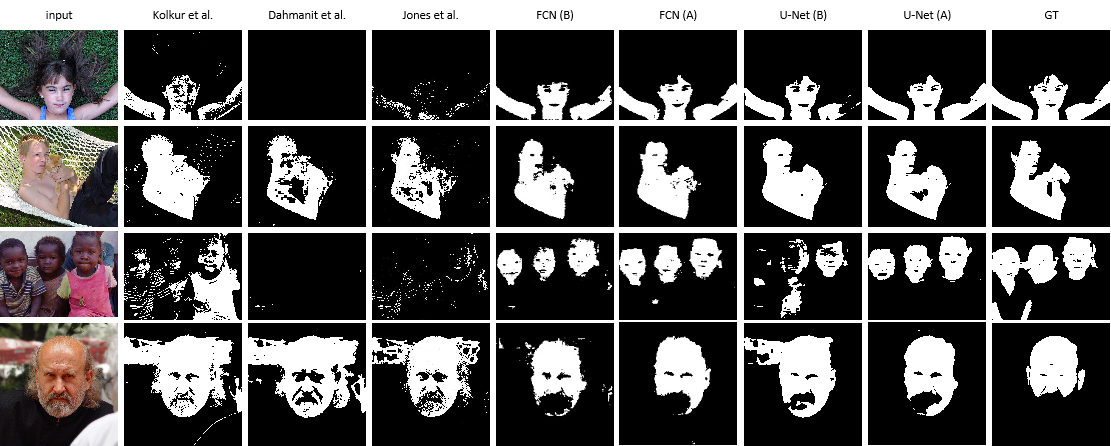}
\vspace{-0.5cm}
\caption{Testing results on the ECU dataset, by various skin segmentation methods including Kolkur~\etal \cite{kolkur2017human}, Dahmani~\etal \cite{Dahmani}, Jones~\etal\cite{jones2002statistical}, FCN before (B) and after (A) augmentation, and U-Net before (B) and after (A) augmentation (Columns 2 to 8).
Input and ground truth are shown in columns 1 and 9.
Our approaches (marked by ``(A)'') achieve superior results for different backgrounds, genders, poses, and skin tones. 
}
\vspace{-0.35cm}
\label{fig:ecu dataset}
\end{figure*}

%%%%%%%%%%%%%%%%%%%%%%%%%%%%%%%%%%%%%%%%%%%%%%%%%%%%%%%%%%%%%%%%%%%%%%%%%%%%%%%%%%%%%%%%%%%%%%%%%%%
In this work, we have used three datasets.
The training and initial evaluation were performed using the benchmark ECU \cite{phung2005skin} dataset.
ECU contains images with diverse attributes including gender, age, skin type, skin-like background, indoor and outdoor images, and images with shadows. The dataset contains 4000 RGB images with manually annotated skin pixels as binary images (see Figure \ref{fig:ecu dataset} as an example). These images are divided into 1600 images for training, 400 for validation, and 2000 for testing. Note that each of these images is used for color space augmentation (18 total). Hence, we have a total of 30400 images for training. To demonstrate the color invariance of the algorithm, we also transformed the test images to augmented space.

%%%%%%%%%%%%%%%%%%%%%%%%%%%%%%%%%%%%%%%%%%%%%%%%%%%%%%%%%%%%%%%%%%%%%%%%%%%%%%%%%%%%%%%%%%%%%%%%%%%

%%%%%%%%%%%%%%%%%%%%%%%%%%%%%%%%%%%%%%%%%%%%%%%%%%%%%%%%%%%%%%%%%%%%%%%%%%%%%%%%%%%%%%%%%%%%%%%%%%%

Racial bias is another critical attribute in skin detection systems. To evaluate such bias in our system, we have experimented with six skin types following the Fitzpatrick scale~\cite{sachdeva2009fitzpatrick}. Figure \ref{fig:ECU skin tones distribution} shows the distribution of images containing individuals with skin types of Type I (less melanin concentration) to Type VI (high concentration of melanin). In the figure, ``mix'' means that  individuals with different skin tones appear in a single image. The figure clearly illustrates the class imbalance within the ECU dataset.
%, which could that can bias any ML algorithm to Type I and Type II. 

\begin{figure*}
\centering
\includegraphics[width=1\linewidth]{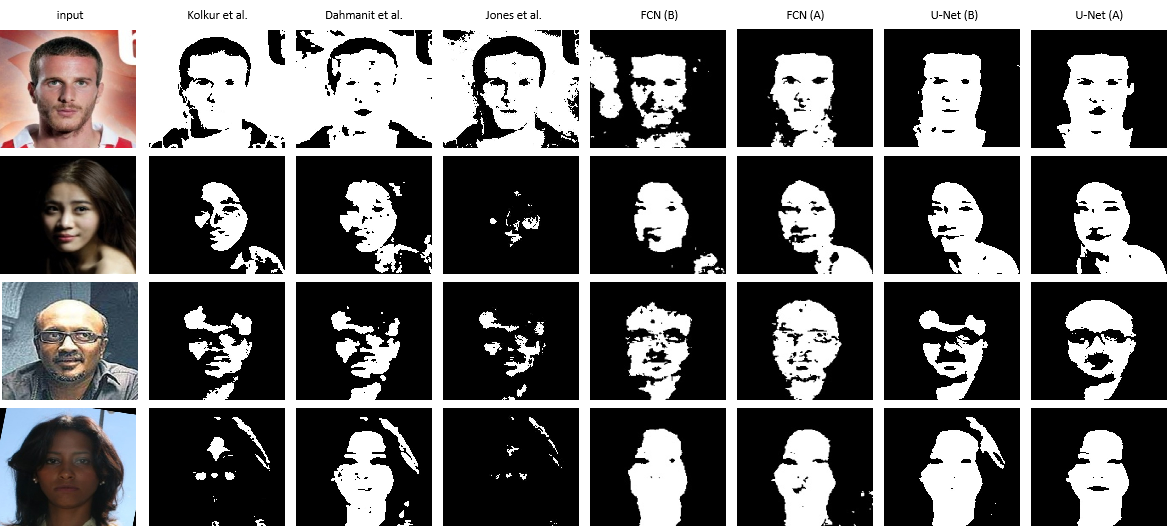}
\vspace{-0.5cm}
\caption{Experimental results from the RFW dataset using several skin segmentation methods. \textit{Left to right:} Kolkur~\etal \cite{kolkur2017human}, Dahmani~\etal\cite{Dahmani}, Jones~\etal\cite{jones2002statistical}, FCN before (B) and after (A) augmentation, and U-Net before (B) and after (A) augmentation.
Rows 1 to 4 show sample results for the RFW ethnic groups: Caucasian, Asian, Indian, and African.
}
\vspace{-0.35cm}
\label{fig:RFW dataset}
\end{figure*}

For further evaluation, we selected the RFW (Racial Faces in the Wild) dataset~\cite{wang2019racial} for \textit{cross dataset validation} of our algorithm in order to test whether the proposed algorithm exhibits bias related to skin tone. %for ethnicity or race. 
RFW is a standard test database used to study racial bias in face recognition (see Figure \ref{fig:RFW dataset} as an example). Four test subsets are provided: Caucasians, Asians, Indians, and Africans. Each subset contains about 3000 individuals and 6000 image pairs for face verification. For our work, the RFW dataset provided 10196 Caucasian faces, 9688 Asian faces, 10308 Indian faces, and 10415 African faces, which shows a good balance across the different groups.

Finally, we created a small dataset of 20 pictures with extreme illumination variations.
These  images were selected to contain either colored neon illumination or artificial filters, as shown in Figure \ref{fig:SkinApparanceVariation}. Then we performed manual annotation using SuperAnnotate \cite{superannotate} to identify the skin pixels and non-skin pixels. We conducted extensive testing using this dataset and made pixel-wise evaluations using our ground-truth annotations. 
%using its  self-annotated ground truth. 
%%%%%%%%%%%%%%%%%%%%%%%%%%%%%%%%%%%%%%%%%%%%%%%%%%%%%%%%%%%%%%%%%%%%%%%%%%%%%%%%%%%%%%%%%%%%%%%%%%%
%%%%%%%%%%%%%%%%%%%%%%%%%%%%%%%%%%%%%%%%%%%%%%%%%%%%%%%%%%%%%%%%%%%%%%%%%%%%%%%%%%%%%%%%%%%%%
\subsection{Evaluation}
%%%%%%%%%%%%%%%%%%%%%%%%%%%%%%%%%%%%%%%%%%%%%%%%%%%%%%%%%%%%%%%%%%%%%%%%%%%%%%%%%%%%%%%%%%%%%
\begin{table}[t]
\small
\begin{center}
\setlength{\tabcolsep}{5pt} % Default value: 6pt
% %%{-.7cm}
\caption{Test results for several skin segmentation methods with the ECU dataset.
Our results using U-Net are significantly better than previous methods. 
For both FCN and U-Net, our use of color-based augmentation improved overall performance of the system.
% \textcolor{blue}{(I tried to write a summary of this table in this caption. Maybe you can provide a stronger summary.)}
}
\label{table:overall evalutations}
\begin{tabular}{c|ccccc}
\hline
Methods & Acc & Pre & Rec & F1 & IoU \\
\hline
Kolkur~\etal \cite{kolkur2017human} &83.73 	&57.00 	&88.38 	&69.31 	&53.03 \\
Dahmani~\etal\cite{Dahmani}  &85.95 	&63.12 	&77.91 	&69.74 	&53.54 \\
Jones~\etal\cite{jones2002statistical} &89.51 	&78.23 	&68.58 	&73.09 	&57.59 \\
FCN before aug. &95.78 	&92.32 	&86.93 	&88.66 	&79.63  \\
FCN after aug. &95.89 	&92.14 	&87.70 	&89.87 	&81.60 \\
U-Net before aug.  &95.59 	&89.56 	&\textbf{89.15} 	&89.35 	&80.76 \\
U-Net after aug.  &\textbf{96.33} 	&\textbf{92.99} 	&89.04 	&\textbf{90.97} 	&\textbf{83.44} \\
\hline
\end{tabular}
\end{center}
\vspace{-0.6cm}
\end{table}

%%%%%%%%%%%%%%%%%%%%%%%%%%%%%%%%%%%%%%%%%%%%%%%%%%%%%%%%%%%%%%%%%%%%%%%%%%%%%%%%%%%%%%%%%%%%%
For the ECU dataset, we used five measures to evaluate the performance: precision, recall, accuracy, F1 score, and IoU. For the RCW dataset, we do not have the ground truth skin annotation. Hence we developed a different method to evaluate the performance. 
We first used a face detector to get the face area in the image. Then we run the skin detection algorithm trained on the ECU dataset. Next, we identify the skin pixels in the face boundary (as shown in Figure \ref{fig:skin in face gt}). Finally, we compute the number of skin pixels in the face boundary to the total number of pixels in the face rectangle. 
%%%%%%%%%%%%%%%%%%%%%%%%%%%%%%%%%%%%%%%%%%%%%%%%%%%%%%%%%%%%%%%%%%%%%%%%%%%%%%%%%%%%%%%%%%%%%%%%%%
\begin{equation}
Skin/Face=\frac{Skin\ pixels\ detected}{Total\ pixels\ in\ face\ rectangle}
\label{fun: skin /face}
\end{equation}
%%{-.2cm}\\
% %%{-.1cm}
%%%%%%%%%%%%%%%%%%%%%%%%%%%%%%%%%%%%%%%%%%%%%%%%%%%%%%%%%%%%%%%%%%%%%%%%%%%%%%%%%%%%%%%%%%%%%%%%%%
As the pose distribution of the image classes in the RCW dataset is uniform, we expect an ideal algorithm to provide a uniform Skin/Face ratio across all ethnicities. 

%%%%%%%%%%%%%%%%%%%%%%%%%%%%%%%%%%%%%%%%%%%%%%%%%%%%%%%%%%%%%%%%%%%%%%%%%%%%%%%%%%%%%%%%%%%%%%%%%%
\begin{table*}
\small
\begin{center}
\setlength{\tabcolsep}{10pt}
\caption{F1 scores (\%) for the ECU dataset across different skin types. The labels I-VI refer to the six skin tones described by Fitzpatrick~\cite{fitzpatrick1988validity}. The  ``mix'' column refers to single images  containing several individuals with  multiple skin  categories. The $\sigma$ column refers to  standard derivation of the F1 scores for all columns. Our method with augmentation outperforms in most skin tone categories.
% \textcolor{blue}{(Here, can we state some important message to take away from this table?)}
}
\vspace{-0.2cm}
\label{table:F1 ECU}
\begin{tabular}{c|cccccccc}
\hline
 Methods& I & II & III & IV & V & VI & mix &$\sigma$\\
\hline
Kolkur~~\etal\cite{kolkur2017human}&67.61 	&69.96 	&70.27 	&70.44 	&67.61 	&46.90 	&72.42 &8.14 \\
Dahmani~~\etal\cite{Dahmani} &66.10 	&70.52 	&71.95 	&71.01 	&70.46 	&56.45 	&70.45 &5.07 \\
Jones~~\etal\cite{jones2002statistical} &64.65 	&75.89 	&73.99 	&74.00 	&73.28 	&46.82 	&77.61 &9.99\\
FCN before aug. &89.03 	&89.90 	&90.03 	&89.56 	&89.59 	&83.41 	&87.37 &2.20\\
FCN after aug. &90.06 	&90.06 	&90.34 	&89.93 	&90.06 	&82.98 	&85.69 &2.70\\
U-Net before aug. &87.16 	&89.58 	&90.38 	&\textbf{90.99} 	&\textbf{91.98} 	&84.72 	&88.82 &2.29\\
U-Net after aug. &\textbf{90.88} &\textbf{91.34} 	&\textbf{91.21} 	&90.55 	&89.35 	&\textbf{86.05} 	&\textbf{89.60} &\textbf{1.84}\\
\hline
\end{tabular}
\end{center}
\vspace{-0.7cm}
\end{table*}
%%%%%%%%%%%%%%%%%%%%%%%%%%%%%%%%%%%%%%%%%%%%%%%%%%%%%%%%%%%%%%%%%%%%%%%%%%%%%%%%%%%%%%%%%%%%%
%%%%%%%%%%%%%%%%%%%%%%%%%%%%%%%%%%%%%%%%%%%%%%%%%%%%%%%%%%%%%%%%%%%%%%%%%%%%%%%%%%%%%%%%%%%%%
\begin{table*}
\small
\begin{center}
\setlength{\tabcolsep}{10pt}
\caption{IoU values (\%) for the ECU dataset across different skin types. The column labels are the same as in the previous table. Our method with augmentation outperforms in most skin tone categories.
% \textcolor{blue}{(Here, can we state an important message to take away from this table?)}
}
\vspace{-0.2cm}
\label{table:IoU ECU}
\begin{tabular}{c|cccccccc}
\hline
Methods & I & II & III & IV & V & VI & mix &$\sigma$\\
\hline
Kolkur~~\etal\cite{kolkur2017human} &51.07 	&53.80 	&54.16 	&54.36 	&51.07 	&30.64 	&56.76 &8.22 \\
Dahmani~~\etal\cite{Dahmani} &49.37 	&54.47 	&56.19 	&55.05 	&54.39 	&39.33 	&54.38  &5.50\\
Jones~~\etal\cite{jones2002statistical} &47.77 	&61.15 	&58.72 	&58.72 	&57.83 	&30.56 	&63.41 &10.60\\
FCN before aug. &80.24 	&81.66 	&81.87 	&81.09 	&81.15 	&71.54 	&77.57 &3.44\\
FCN after aug. &81.92 	&81.91 	&82.38 	&81.70 	&81.91 	&70.92 	&74.97 &4.22\\
U-Net before aug. &77.24 	&81.13 	&82.44 	&\textbf{83.47} 	&\textbf{85.16} 	&73.50 	&79.89 &3.68\\
U-Net after aug. &\textbf{83.28} 	&\textbf{84.06} 	&\textbf{83.84} 	&82.73 	&80.75 	&\textbf{75.52} 	&\textbf{81.16} &\textbf{2.97}\\
\hline
\end{tabular}
\end{center}
\vspace{-0.8cm}
\end{table*}

% %%%%%%%%%%%%%%%%%%%%%%%%%%%%%%%%%%%%%%%%%%%%%%%%%%%%%%%%%%%%%%%%%%%%%%%%%%%%%%%%%%%%%%%%%%%

\section{Results and Discussion}
\subsection{Images in the wild}
% \textcolor{blue}{(It seems that this section should also mention the FCN results that are in the table.)}
We compared our method with some state-of-the-art skin segmentation systems, including three traditional methods and one FCN based methods. Kolkur~\etal\cite{kolkur2017human} and Dahmani~\etal\cite{Dahmani} are two thresholding methods which establish some rules in several color spaces to classify a pixel is skin or not. Jones~\etal\cite{jones2002statistical} is a naive bayes based methods, which predicts the probability of a pixel to be skin after training with given skin masks. 
The problem behind these traditional methods is the lack of high level features during detection tasks, resulting in the weak robustness against light changes, complex backgrounds or skin color diversity.
For both FCN based methods and our U-Net based method, we trained two models, one without color augmentation and another with color augmentation to confirm the effectiveness of color augmentation.

We first trained the U-Net model with the original RGB images in the ECU dataset (with and without augmentation) and evaluated the performance with the original test set. The precision and recall are shown in Table \ref{table:overall evalutations}. With augmentation, this system yielded a precision of 92.99\% and recall of 89.04\%, which significantly outperforms the methods of Kolkur~\etal \cite{kolkur2017human}, Dahmani~\etal \cite{Dahmani}, and Jones~\etal \cite{jones2002statistical} (Naive Bayes). The FCN model achieves a precision of 92.14\% and a recall of 87.70\%.
This model also outperforms most  CNN-based methods in terms of overall accuracy. While our method shows an accuracy of 96.33\%, Tarasiewicz~\cite{tarasiewicz2020skinny} (also a U-Net based architecture) reported an accuracy of 92\%.
% , and Zuo~\etal \cite{zuo2017combining} (RNN along with FCN) reported an accuracy of 98.10\%. 
% \textcolor{blue}{(Can we say why Zuo et al. were better? Or describe some limitation of their system?)} 
% Topiwala~\etal\cite{8941944} (U-Net) reported an accuracy of 95.51\% on a custom dataset that only targets abdomen images.

In Figure \ref{fig:ecu dataset} we show qualitative comparisons, where the examples cover various skin colors, similar colors in the background, and complex illumination.
The first row is a girl lying on the grass with her arms open. The second row is a boy holding a cat, and some of his skin areas are covered by shadow and the cat. The third row is an image of three children with dark skin. The fourth row is a man with a large beard area on his face. These challenging conditions make other methods fail or perform poorly. Dahmani~\etal\cite{Dahmani} and Jones~\etal\cite{jones2002statistical} fail in the first and third rows. Kolkur~\etal\cite{kolkur2017human} classified a large area of background as skin pixels in the fourth row. U-Net (B) works better but still performs poorly in the third row. In contrast, our approach overcame most of the difficulties as stated above, and produced accurate and robust results.
(More examples are provided in the supplementary material.)

In order to detect the skin tone bias in the ECU dataset, we further tested the algorithms on different skin tones. 
Table \ref{table:F1 ECU} and Table \ref{table:IoU ECU} show that our method outperforms in all categories. 
Among the three baseline methods,  Jones\cite{jones2002statistical} shows the best performance for most of the skin types, but all the methods particularly fail for Type VI (dark skin category). Even deep learning skin segmentation methods show an apparent decline in this dark skin category. 
Our method consistently shows more than 85\% F1 score and more than 75\% IoU for all skin types. Moreover, the standard derivations in the last column show that deep learning models have more substantial stability over skin tone bias after color augmentation (more details in the supplementary material).

As shown in Figure \ref{fig:ECU skin tones distribution}, one of the significant problems in the ECU dataset is the imbalance in the images for each skin type. To further understand the robustness of our method, we test with the RFW dataset, which has a balanced dataset across ethnicity. 
Since RFW dataset does contain manual labels, we only compute the skin/face ratio in this part (see Figure \ref{fig:skin /face} and Function \ref{fun: skin /face}).
Table \ref{table:skin/face rfw} shows the evaluated skin/face result  of RFW dataset. The model is trained on the ECU dataset using data augmentation. 
The results show that other methods have different degrees of decline in the ``African'' group, while our method is stable among different races. Our method outperforms the best method for the African group by nearly 29\%. Also, our method is significantly better in all the categories compared to the three baselines.
Compared the results shown in the last four rows, color augmentation shows its effectiveness on improving the performance of the models. Considering the problems of over-prediction, more discussions are shown in the supplementary materials.\par
\begin{table}
\small
% \vspace{-0.2cm}
\begin{center}
\caption{Skin/face ratios (\%) for the four ethnic groups of the RFW dataset.
For the first 3 methods, a significant decline is present for the ``African'' group.
That decline is not present for our FCN and U-Net models, after training with color-space augmentation.
}
% %%{-.cm}
\label{table:skin/face rfw}
\setlength{\tabcolsep}{1.5mm}{
\begin{tabular}{c|ccccc}
\hline
Methods& Cau & Asian & Ind & Afr\\% &$\sigma$\\
\hline
Kolkur~~\etal\cite{kolkur2017human} &62.34 	&62.21 	&64.31 	&36.78 \\
 %&11.36\\
Dahmani~\etal~\cite{Dahmani} &60.02 	&59.19 	&60.95 	&49.79 \\
 %&4.49\\
Jones~~\etal\cite{jones2002statistical} &47.29 	&45.49 	&48.45 	&20.05 \\
 %&11.75\\
FCN before aug. &67.24 	&65.78 	&67.36 	&64.37 \\
% &\textbf{1.22} \\
FCN after aug. &73.35 	&70.59 	&72.73 	&73.99 \\
 %&1.28\\
U-Net before aug. & 65.47 	&65.76 	&69.82 	&68.39 \\
 %&1.82\\
U-Net after aug. & \textbf{77.17} & \textbf{73.20} &\textbf{76.16} &\textbf{78.73}\\ %&2.34\\
\hline
\end{tabular}}
\end{center}
\vspace{-0.7cm}
\end{table}

\begin{figure*}[t]
\centering
\begin{subfigure}{0.3\textwidth}
\includegraphics[width=1.0\linewidth]{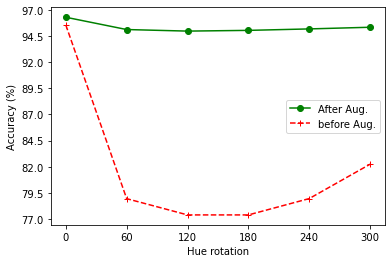} 
\caption{}
\label{fig:hue change}
\end{subfigure}
\quad
\begin{subfigure}{0.3\textwidth}
\includegraphics[width=1.0\linewidth]{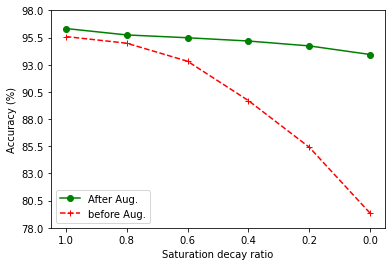}
\caption{}
\label{fig:saturation change}
\end{subfigure}
\quad
\begin{subfigure}{0.3\textwidth}
\includegraphics[width=1.0\linewidth]{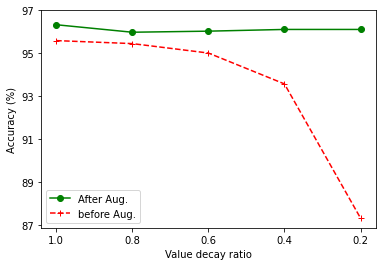}
\caption{}
\label{fig:value change}
\end{subfigure}
\vspace{-0.35cm}
\caption{Comparison of U-Net models using color augmentation (``After Aug.'') and without color augmentation (``Before Aug.'') to test robustness for image filtering in the color space. Augmentation shows their effectiveness for all three dimensions: (a) hue, (b) saturation, and (c) value. 
}
\vspace{-0.4cm}
\label{fig:testing accuracy change HSV}
\end{figure*}
Figure \ref{fig:RFW dataset} shows some qualitative results containing both various skin colors and complex illuminance.
The first row is a Caucasian man with background of  color similar to his skin. The second row is an Asian woman with one shoulders in the dark. The third row is an Indian man with strong light on his top head. The fourth row is an African woman with her face in shadow. 
We find that the three baseline methods in columns 2 to 4 are fully confused by the background in the first row. U-Net (B) fails to detect the skin area with intense light in the fourth row. In contrast, our method works better and outputs accurate and complete results.

%%%%%%%%%%%%%%%%%%%%%%%%%%%%%%%%%%%%%%%%%%%%%%%%%%%%%%%%%%%%%%%%%%%%%%%%%%%%%%%%%%%%%%%%%%%%%%%%%%%

% %%{-.3cm}
%%%%%%%%%%%%%%%%%%%%%%%%%%%%%%%%%%%%%%%%%%%%%%%%%%%%%%%%%%%%%%%%%%%%%%%%%%%%%%%%%%%%%%%%%%%%%%%%%%%

\subsection{Cross dataset testing and ablation study}
%%%%%%%%%%%%%%%%%%%%%%%%%%%%%%%%%%%%%%%%%%%%%%%%%%%%%%%%%%%%%%%%%%%%%%%%%%%%%%%%%%%%%%%%%%%%%%%%%%%

%%%%%%%%%%%%%%%%%%%%%%%%%%%%%%%%%%%%%%%%%%%%%%%%%%%%%%%%%%%%%%%%%%%%%%%%%%%%%%%%%%%%%%%%%%%%%%%%%%%
In order to further test the robustness of the algorithm under different illumination and creative filtering conditions, we ran experiments by testing images transformed by HSV color space augmentation (similar to the training set). We trained two skin segmentation algorithms: One without color augmentation and one with color augmentation. We tested them with the color augmented test set. Figure \ref{fig:testing accuracy change HSV} shows the comparison between the results. For the model without color augmentation, the accuracy falls immediately when tested with images that are modified by hue (Figure \ref{fig:hue change}), saturation (Figure \ref{fig:saturation change}), or value (Figure \ref{fig:value change}). The accuracy remains consistent for the model that was trained with a set of color augmented images. This ablation study shows explicitly the effectiveness of our proposed model. 

Figure \ref{fig:HSV augmentation and result} shows qualitative examples of how the output of our algorithm remains consistent across all the filters, even with drastic changes in the color. Finally, in order to test the performance under ambient light across the spectral, we selected random images from the web and tested them. 

Figure \ref{fig:SkinApparanceVariation} shows the robustness of our methods against the drastic illumination changes. 
The model without color augmentation fails to detect a single skin pixel in the second and third rows, while our method (with augmentation) successfully detects skin pixels in most cases. 
% \textcolor{blue}{(The previous sentence is not true, is it?)}
Qualitative evaluations in Table \ref{table:illumination evaluation} shows this improvement. IoU increase sharply after color augmentation is applied to the system. More results are in the supplementary material.

\begin{table}
\small
\begin{center}
% \vspace{-0.5cm}
\caption{
Augmentation improves the performance of both U-Net and FCN when tested on images with unconstrained illumination and filters from our self-made dataset. 
}
\vspace{-0.35cm}
\label{table:illumination evaluation}
%%{-.2cm}
\begin{tabular}{c|ccc}
\hline
 & IoU before aug. & IoU after aug. & IoU gain by aug. \\
\hline
FCN  	&12.05 &64.92 &$\uparrow$ 52.87 \\
U-Net  &11.76 &35.85 &$\uparrow$ 24.09	\\
\hline
\end{tabular}
\end{center}
\vspace{-0.5cm}
\end{table}
% %%{-.7cm}

\begin{table}
\small
\begin{center}
% \vspace{-0.5cm}
\caption{
Augmentation improves the performance of both U-Net and FCN when tested on grayscale images from the ECU dataset.
}
\vspace{-0.35cm}
\label{table:grayscale evaluation}
%%{-.2cm}
\begin{tabular}{c|ccc}
\hline
 & IoU before aug. & IoU after aug. & IoU gain by aug. \\
\hline
FCN  	&47.13 &77.20 &$\uparrow$ 30.07 \\
U-Net  &0.55 &69.42 &$\uparrow$ 68.87	\\
\hline
\end{tabular}
\end{center}
\vspace{-0.75cm}
\end{table}

\subsubsection{Grayscale images}
We also conducted tests  using grayscale images only. In this part, we used images from the ECU dataset and converted them to grayscale format. We performed testing with the U-Net model and the FCN model, with and without color space augmentation. The results are listed in Table \ref{table:grayscale evaluation}.

The table shows that the FCN model without color-space augmentation yields very poor performance,  and the resulting IoU is low. For U-Net without augmentation, IoU declines to approximately 0, indicating that the model detected hardly any skin pixels.
The performance improves significantly for both of these models, when color augmentation was applied. IoU for the FCN model  returned to nearly 80\%. 
Improvements also happen to the 
IoU for the U-Net model rose to nearly 70\% with the help of color augmentation. 
We list some example results in the Figure \ref{fig:SkinApparanceVariation} and more results can be found in the supplementary material.

\section{Conclusion}
This paper has introduced a new approach for automated detection of skin in images. 
% The system leverages recent innovations involving fully convolutional networks and encoder-decoder networks like U-Net.
Experimental results show that the color-based data augmentation step strategically reduces dependence by the system on color-based cues, and thereby reduces %racial bias. 
bias related to lightness and color of skin.
% The approach addresses problems related to illumination differences (e.g., indoor/outdoor situations, harsh shadows, unnatural lighting), variations in skin tone (especially ethnic/racial variations), and different sensor parameters (e.g., color, monochromatic, varying spectral sensitivities). 
% To our knowledge, the system presented here is the first to apply deep methods to the problem of skin detection that can be generalized across the color spectrum.
% Further, this work is the first to directly assess  ethnic or racial bias in skin detection systems. 
 % are presented using two datasets.
Using the ECU dataset, our approach has demonstrated better performance than three alternative skin-detection systems.
%Our approach has demonstrated better performance than three alternative skin-detection systems for the ECU dataset. 
We also conducted an experiment using a racially-balanced dataset,  RFW, to illustrate the robustness of our method across different skin tones. %against  skin tones changes.
Further, the approach addresses problems related to illumination differences (e.g., indoor/outdoor situations, harsh shadows, unnatural lighting) and different sensor parameters (e.g., color, monochromatic, varying spectral sensitivities). 
We anticipate that similar approaches can be applied more broadly to other Computer Vision tasks.

%%%%%%%%% REFERENCES
{\small
\bibliographystyle{ieee_fullname}
\bibliography{PaperForReview}

\begin{thebibliography}{10}\itemsep=-1pt

\bibitem{adeli2021representation}
Ehsan Adeli, Qingyu Zhao, Adolf Pfefferbaum, Edith~V. Sullivan, Li Fei-Fei,
  Juan~Carlos Niebles, and Kilian~M. Pohl.
\newblock Representation learning with statistical independence to mitigate
  bias.
\newblock In {\em Proceedings of the IEEE/CVF Winter Conference on Applications
  of Computer Vision}, pages 2513--2523, 2021.

\bibitem{alsheakhali2011hand}
Mohamed Alsheakhali, Ahmed Skaik, Mohammed Aldahdouh, and Mahmoud Alhelou.
\newblock Hand gesture recognition system.
\newblock {\em Information \& Communication Systems}, 132--136, 2011.

\bibitem{badrinarayanan2017segnet}
Vijay Badrinarayanan, Alex Kendall, and Roberto Cipolla.
\newblock {SegNet}: A deep convolutional encoder-decoder architecture for image
  segmentation.
\newblock {\em IEEE Transactions on Pattern Analysis and Machine Intelligence},
  39(12):2481--2495, 2017.

\bibitem{borshukov2005realistic}
George Borshukov and John~P. Lewis.
\newblock Realistic human face rendering for `{The Matrix Reloaded}'.
\newblock In {\em ACM Siggraph 2005 Courses}, pages 13--es. 2005.

\bibitem{buolamwini2018gender}
Joy Buolamwini and Timnit Gebru.
\newblock Gender shades: Intersectional accuracy disparities in commercial
  gender classification.
\newblock In {\em Conference on Fairness, Accountability and Transparency},
  pages 77--91. PMLR, 2018.

\bibitem{buolamwini2017gender}
Joy~Adowaa Buolamwini.
\newblock {\em Gender shades: intersectional phenotypic and demographic
  evaluation of face datasets and gender classifiers}.
\newblock PhD thesis, Massachusetts Institute of Technology, 2017.

\bibitem{cula2005skin}
Oana~G. Cula, Kristin~J. Dana, Frank~P. Murphy, and Babar~K. Rao.
\newblock Skin texture modeling.
\newblock {\em International Journal of Computer Vision}, 62(1):97--119, 2005.

\bibitem{Dahmani}
Djamila Dahmani, Mehdi Cheref, and Slimane Larabi.
\newblock Zero-sum game theory model for segmenting skin regions.
\newblock {\em Image and Vision Computing}, 99:103925, 2020.

\bibitem{opensourseFCN}
Yunlong Dong.
\newblock Trying to be the easiest {FCN} {P}y{T}orch implementation.
\newblock \url{https://github.com/yunlongdong/FCN-pytorch}.

\bibitem{donner2008layered}
Craig Donner, Tim Weyrich, Eugene d'Eon, Ravi Ramamoorthi, and Szymon
  Rusinkiewicz.
\newblock A layered, heterogeneous reflectance model for acquiring and
  rendering human skin.
\newblock {\em ACM Transactions on Graphics (TOG)}, 27(5):1--12, 2008.

\bibitem{drimbarean2001image}
Alexandru~F. Drimbarean, Peter~M. Corcoran, Mihai Cuic, and Vasile Buzuloiu.
\newblock Image processing techniques to detect and filter objectionable images
  based on skin tone and shape recognition.
\newblock In {\em Proceedings of the International Conference on Consumer
  Electronics}, pages 278--279. IEEE, 2001.

\bibitem{drozdowski2020demographic}
Pawel Drozdowski, Christian Rathgeb, Antitza Dantcheva, Naser Damer, and
  Christoph Busch.
\newblock Demographic bias in biometrics: A survey on an emerging challenge.
\newblock {\em IEEE Transactions on Technology and Society}, 1(2):89--103,
  2020.

\bibitem{fitzpatrick1988validity}
Thomas~B. Fitzpatrick.
\newblock The validity and practicality of sun-reactive skin types {I} through
  {VI}.
\newblock {\em Archives of Dermatology}, 124(6):869--871, 1988.

\bibitem{forsyth1996finding}
David~A. Forsyth, Margaret Fleck, and Chris Bregler.
\newblock Finding naked people.
\newblock {\em International Journal of Computer Vision}, 1065(1):593--602,
  1996.

\bibitem{fotouhi2009skin}
Mehran Fotouhi, Mohammad~H. Rohban, and Shohreh Kasaei.
\newblock Skin detection using contourlet-based texture analysis.
\newblock In {\em Proceedings of the Fourth International Conference on Digital
  Telecommunications}, pages 59--64. IEEE, 2009.

\bibitem{garcia1999face}
Christophe Garcia and George Tziritas.
\newblock Face detection using quantized skin color regions merging and wavelet
  packet analysis.
\newblock {\em IEEE Transactions on Multimedia}, 1(3):264--277, 1999.

\bibitem{girshick2014rich}
Ross Girshick, Jeff Donahue, Trevor Darrell, and Jitendra Malik.
\newblock Rich feature hierarchies for accurate object detection and semantic
  segmentation.
\newblock In {\em Proceedings of the IEEE Conference on Computer Vision and
  Pattern Recognition}, pages 580--587, 2014.

\bibitem{he2017mask}
Kaiming He, Georgia Gkioxari, Piotr Doll{\'a}r, and Ross Girshick.
\newblock Mask {R-CNN}.
\newblock In {\em Proceedings of the IEEE International Conference on Computer
  Vision}, pages 2961--2969, 2017.

\bibitem{he2019semi}
Yi He, Jiayuan Shi, Chuan Wang, Haibin Huang, Jiaming Liu, Guanbin Li, Risheng
  Liu, and Jue Wang.
\newblock Semi-supervised skin detection by network with mutual guidance.
\newblock In {\em Proceedings of the IEEE/CVF International Conference on
  Computer Vision}, pages 2111--2120, 2019.

\bibitem{igarashi2005appearance}
Takanori Igarashi, Ko Nishino, and Shree~K. Nayar.
\newblock The appearance of human skin.
\newblock Department of Computer Science, Columbia University, 2005.

\bibitem{jones2002statistical}
Michael~J. Jones and James~M. Rehg.
\newblock Statistical color models with application to skin detection.
\newblock {\em International Journal of Computer Vision}, 46(1):81--96, 2002.

\bibitem{kakumanu2007survey}
Praveen Kakumanu, Sokratis Makrogiannis, and Nikolaos Bourbakis.
\newblock A survey of skin-color modeling and detection methods.
\newblock {\em Pattern Recognition}, 40(3):1106--1122, 2007.

\bibitem{keras}
Keras.
\newblock Image data preprocessing.
\newblock \url{https://keras.io/api/preprocessing/image/}.

\bibitem{kolkur2017human}
S. Kolkur, D. Kalbande, P. Shimpi, C. Bapat, and J. Jatakia.
\newblock Human skin detection using {RGB}, {HSV} and {YCbCr} color models.
\newblock In {\em Proceedings of the International Conference on Communication
  and Signal Processing}, pages 324--332. Atlantis Press, 2016/12.

\bibitem{kovac2003human}
J. {Kovac}, P. {Peer}, and F. {Solina}.
\newblock Human skin color clustering for face detection.
\newblock In {\em Proceedings of The IEEE Region 8 EUROCON 2003. Computer as a
  Tool}, volume~2, pages 144--148, 2003.

\bibitem{li2019repair}
Yi Li and Nuno Vasconcelos.
\newblock Repair: Removing representation bias by dataset resampling.
\newblock In {\em Proceedings of the IEEE/CVF Conference on Computer Vision and
  Pattern Recognition}, pages 9572--9581, 2019.

\bibitem{liu2010robust}
Qiong Liu and Guang-zheng Peng.
\newblock A robust skin color based face detection algorithm.
\newblock In {\em Proceedings of the 2nd International Asia Conference on
  Informatics in Control, Automation and Robotics (CAR 2010)}, volume~2, pages
  525--528. IEEE, 2010.

\bibitem{long2015fully}
Jonathan Long, Evan Shelhamer, and Trevor Darrell.
\newblock Fully convolutional networks for semantic segmentation.
\newblock In {\em Proceedings of the IEEE Conference on Computer Vision and
  Pattern Recognition}, pages 3431--3440, 2015.

\bibitem{mahmoodi2016comprehensive}
Mohammad~Reza Mahmoodi and Sayed~Masoud Sayedi.
\newblock A comprehensive survey on human skin detection.
\newblock {\em International Journal of Image, Graphics and Signal Processing},
  8(5):1, 2016.

\bibitem{mehrabi2021survey}
Ninareh Mehrabi, Fred Morstatter, Nripsuta Saxena, Kristina Lerman, and Aram
  Galstyan.
\newblock A survey on bias and fairness in machine learning.
\newblock {\em ACM Computing Surveys (CSUR)}, 54(6):1--35, 2021.

\bibitem{nguyen2019hand}
Duong~Hai Nguyen, Tai~Nhu Do, In-Seop Na, and Soo-Hyung Kim.
\newblock Hand segmentation and fingertip tracking from depth camera images
  using deep convolutional neural network and multi-task {SegNet}.
\newblock {\em arXiv preprint arXiv:1901.03465}, 2019.

\bibitem{nguyen2018hand}
Dinh-Ha Nguyen, Trung-Hieu Le, Thanh-Hai Tran, Hai Vu, Thi-Lan Le, and
  Huong-Giang Doan.
\newblock Hand segmentation under different viewpoints by combination of {Mask
  R-CNN} with tracking.
\newblock In {\em Proceedings of the 5th Asian Conference on Defense Technology
  (ACDT)}, pages 14--20. IEEE, 2018.

\bibitem{phung2005skin}
Son~Lam Phung, Abdesselam Bouzerdoum, and Douglas Chai.
\newblock Skin segmentation using color pixel classification: analysis and
  comparison.
\newblock {\em IEEE Transactions on Pattern Analysis and Machine Intelligence},
  27(1):148--154, 2005.

\bibitem{ronneberger2015u}
Olaf Ronneberger, Philipp Fischer, and Thomas Brox.
\newblock {U-Net}: Convolutional networks for biomedical image segmentation.
\newblock In {\em International Conference on Medical Image Computing and
  Computer-assisted Intervention}, pages 234--241. Springer, 2015.

\bibitem{roy2017deep}
Kankana Roy, Aparna Mohanty, and Rajiv~R. Sahay.
\newblock Deep learning based hand detection in cluttered environment using
  skin segmentation.
\newblock In {\em Proceedings of the IEEE International Conference on Computer
  Vision Workshops}, pages 640--649, 2017.

\bibitem{sachdeva2009fitzpatrick}
Silonie Sachdeva.
\newblock Fitzpatrick skin typing: Applications in dermatology.
\newblock {\em Indian Journal of Dermatology, Venereology and Leprology},
  75(1):93, 2009.

\bibitem{salimi2019interventional}
Babak Salimi, Luke Rodriguez, Bill Howe, and Dan Suciu.
\newblock Interventional fairness: Causal database repair for algorithmic
  fairness.
\newblock In {\em Proceedings of the 2019 International Conference on
  Management of Data}, pages 793--810, 2019.

\bibitem{sarkar2017universal}
Abhijit Sarkar, A.~Lynn Abbott, and Zachary Doerzaph.
\newblock Universal skin detection without color information.
\newblock In {\em Proceedings of the IEEE Winter Conference on Applications of
  Computer Vision (WACV)}, pages 20--28, 2017.

\bibitem{sixta2020fairface}
Tom{\'a}{\v{s}} Sixta, Julio C. S.~Jacques Junior, Pau Buch-Cardona, Eduard
  Vazquez, and Sergio Escalera.
\newblock {FairFace} challenge at {ECCV} 2020: Analyzing bias in face
  recognition.
\newblock In {\em Proceedings of the European Conference on Computer Vision},
  pages 463--481. Springer, 2020.

\bibitem{superannotate}
SuperAnnotate.
\newblock The ultimate training data platform for {AI}.
\newblock \url{https://www.superannotate.com/}.

\bibitem{tarasiewicz2020skinny}
Tomasz Tarasiewicz, Jakub Nalepa, and Michal Kawulok.
\newblock Skinny: A lightweight {U-Net} for skin detection and segmentation.
\newblock In {\em Proceedings of the International Conference on Image
  Processing (ICIP)}, pages 2386--2390. IEEE, 2020.

\bibitem{8941944}
A. {Topiwala}, L. {Al-Zogbi}, T. {Fleiter}, and A. {Krieger}.
\newblock Adaptation and evaluation of deep learning techniques for skin
  segmentation on novel abdominal dataset.
\newblock In {\em Proceedings of the 19th International Conference on
  Bioinformatics and Bioengineering (BIBE)}, pages 752--759, 2019.

\bibitem{vezhnevets2003survey}
Vladimir Vezhnevets, Vassili Sazonov, and Alla Andreeva.
\newblock A survey on pixel-based skin color detection techniques.
\newblock In {\em Proc. Graphicon}, volume~3, pages 85--92. Citeseer, 2003.

\bibitem{wang2019racial}
Mei Wang, Weihong Deng, Jiani Hu, Xunqiang Tao, and Yaohai Huang.
\newblock Racial faces in the wild: Reducing racial bias by information
  maximization adaptation network.
\newblock In {\em Proceedings of the IEEE/CVF International Conference on
  Computer Vision}, pages 692--702, 2019.

\bibitem{opensourseUNet}
Xuhao Zhi.
\newblock Implementation of deep learning framework -- {U-Net}, using {Keras}.
\newblock \url{https://github.com/zhixuhao/unet}.

\bibitem{zollhofer2018state}
Michael Zollh{\"o}fer, Justus Thies, Pablo Garrido, Derek Bradley, Thabo
  Beeler, Patrick P{\'e}rez, Marc Stamminger, Matthias Nie{\ss}ner, and
  Christian Theobalt.
\newblock State of the art on monocular {3D} face reconstruction, tracking, and
  applications.
\newblock In {\em Computer Graphics Forum}, volume~37, pages 523--550. Wiley
  Online Library, 2018.

\bibitem{zuo2017combining}
Haiqiang Zuo, Heng Fan, Erik Blasch, and Haibin Ling.
\newblock Combining convolutional and recurrent neural networks for human skin
  detection.
\newblock {\em IEEE Signal Processing Letters}, 24(3):289--293, 2017.

\end{thebibliography}


\begin{thebibliography}{1}\itemsep=-1pt

\bibitem{Dahmani}
Djamila Dahmani, Mehdi Cheref, and Slimane Larabi.
\newblock Zero-sum game theory model for segmenting skin regions.
\newblock {\em Image and Vision Computing}, 99:103925, 2020.

\bibitem{jones2002statistical}
Michael~J. Jones and James~M. Rehg.
\newblock Statistical color models with application to skin detection.
\newblock {\em International Journal of Computer Vision}, 46(1):81--96, 2002.

\bibitem{kolkur2017human}
S. Kolkur, D. Kalbande, P. Shimpi, C. Bapat, and J. Jatakia.
\newblock Human skin detection using {RGB}, {HSV} and {YCbCr} color models.
\newblock In {\em Proceedings of the International Conference on Communication
  and Signal Processing}, pages 324--332. Atlantis Press, 2016/12.

\bibitem{Qin_2019_CVPR}
Xuebin Qin, Zichen Zhang, Chenyang Huang, Chao Gao, Masood Dehghan, and Martin
  Jagersand.
\newblock Basnet: Boundary-aware salient object detection.
\newblock In {\em Proceedings of the IEEE/CVF Conference on Computer Vision and
  Pattern Recognition (CVPR)}, June 2019.

\end{thebibliography}
}

\end{document}

%%%%%%%%% TITLE - PLEASE UPDATE
\title{Supplementary Material}

% \author{First Author\\
% Institution1\\
% Institution1 address\\
% {\tt\small firstauthor@i1.org}
% For a paper whose authors are all at the same institution,
% omit the following lines up until the closing ``}''.
% Additional authors and addresses can be added with ``\and'',
% just like the second author.
% To save space, use either the email address or home page, not both
% \and
% Second Author\\
% Institution2\\
% First line of institution2 address\\
% {\tt\small secondauthor@i2.org}
% }
\maketitle

%%%%%%%%% ABSTRACT
% \begin{abstract}
%   The ABSTRACT is to be in fully justified italicized text, at the top of the left-hand column, below the author and affiliation information.
%   Use the word ``Abstract'' as the title, in 12-point Times, boldface type, centered relative to the column, initially capitalized.
%   The abstract is to be in 10-point, single-spaced type.
%   Leave two blank lines after the Abstract, then begin the main text.
%   Look at previous CVPR abstracts to get a feel for style and length.
% \end{abstract}

%%%%%%%%% BODY TEXT
% \section{More training details}
\section{Results on the ECU dataset}
\label{sec:intro}
\subsection{Quantitative results}
\begin{figure}
\begin{center}
  \includegraphics[width=1\linewidth]{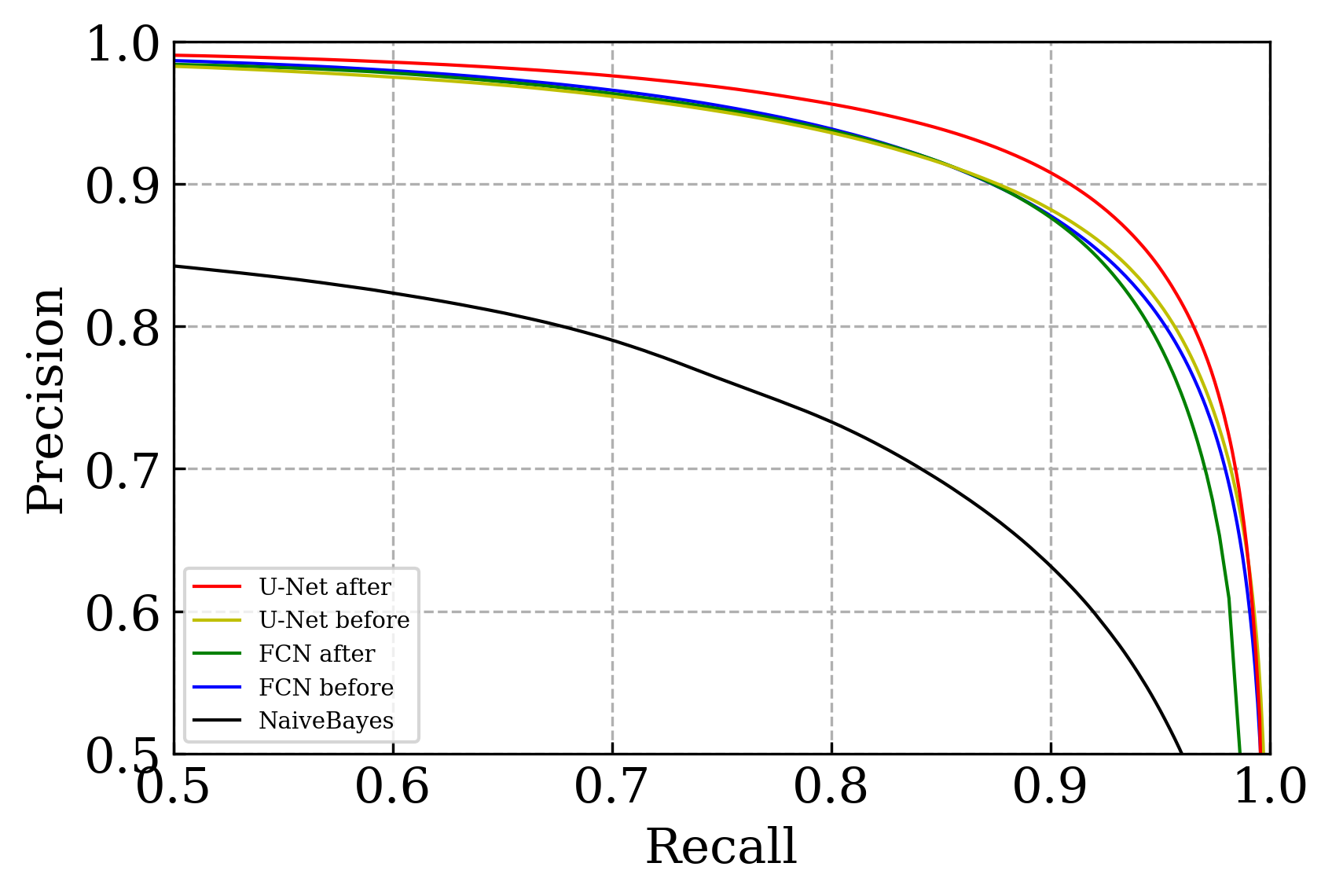}
\end{center}
  \caption{Precision-recall curve from testing experiments on the ECU dataset.}
\label{fig:p-r curve}
\end{figure}

The main paper set the threshold of predicted probability to be 0.5 to classify the image pixels as skin or non-skin pixels and convert the sigmoid output from the deep learning model to binary. To make the experimental testing results more convincing, we illustrate the precision-recall curve\cite{Qin_2019_CVPR} in Figure \ref{fig:p-r curve}.

\subsection{Qualitative results}
\begin{figure*}
\begin{center}
  \includegraphics[width=1\linewidth]{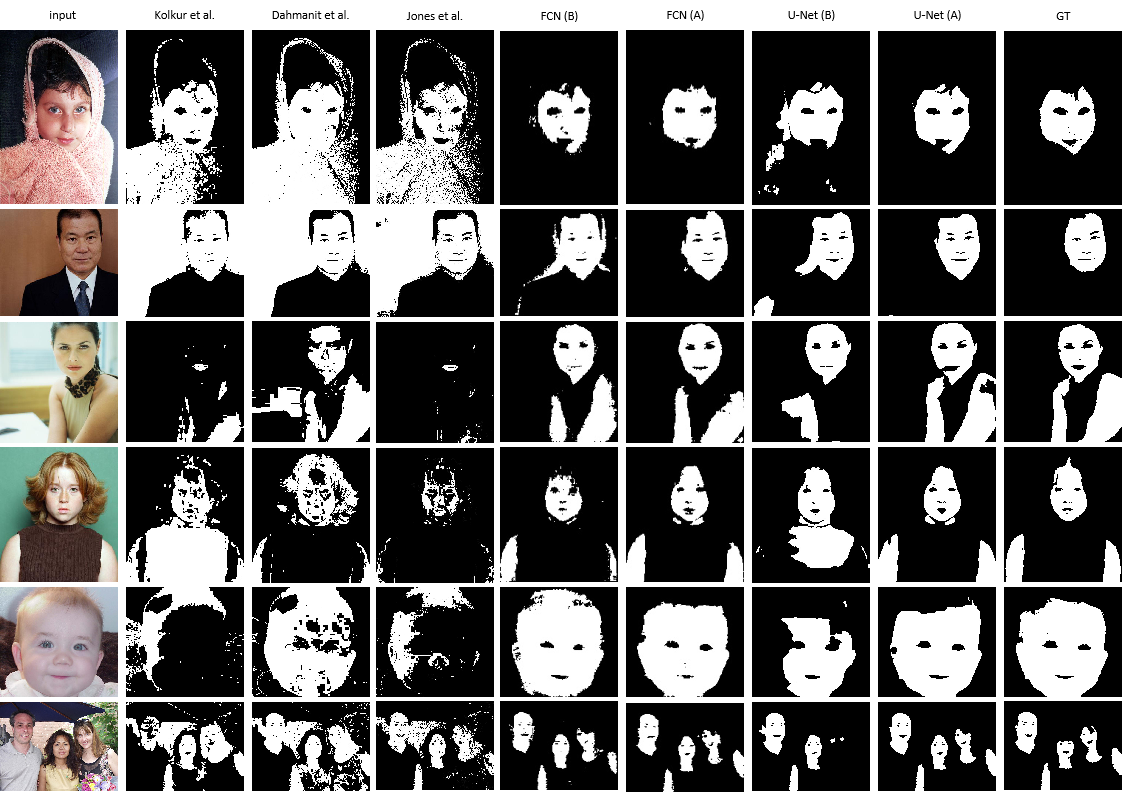}
\end{center}
  \caption{Additional results on the ECU dataset, by various skin segmentation methods including Kolkur ~\etal \cite{kolkur2017human}, Dahmani~\etal\cite{Dahmani}, Jones~\etal\cite{jones2002statistical}, FCN before (B) and after (A) augmentation, and U-Net before (B) and after (A) augmentation (Columns 2 to 8).
Input and ground truth are shown in column 1 and 9.}
\label{fig:more results ecu}
\end{figure*}

This section will supply more output results from the skin detection systems mentioned in the main paper. 
We illustrate another six examples in Figure \ref{fig:more results ecu}. 
The first row is a girl wearing a skin color-like cloth. 
The second and third rows contain backgrounds that have similar colors as the people in the image. 
The fourth row is a girl with brown cloth.
The fifth row is a baby with strong lights on his head.
The sixth row contains multiple people in various poses and skin colors
These challenging conditions make other methods fail or perform poorly. 
The three baseline methods all fail to classify the skin color-like ground in the first and second rows.
U-Net (B) works better, but there is still some false positive noise in rows 1 to 5. Moreover, it fails to detect the people on the right in the last row.
In contrast, our approach overcame most of the difficulties mentioned above and produced accurate and robust results.
Compared with results before color augmentation, models with color augmentation make less false positive and false negative judgments. For example, the FCN (A) does not detect the baby's hair as skin pixels in the fifth row, and it does not make noise as FCN (B) does in the second row.
In this part of view, our method outperforms the traditional skin segmentation methods, and color augmentation helps deep learning methods work better. 

%%%%%%%%%%%%%%%%%%%%%%%%%%%%%%%%%%%%%%%%%%%%%%%%%%%%%%%%%%%%%%%%%%%%%%%%%%%%%%%%%%%%%%%%%%%%%%%%%%%%%%%%%%
% %%%%%%%%%%%%%%%%%%%%%%%%%%%%%%%%%%%%%%%%%%%%%%%%%%%%%%%%%%%%%%%%%%%%%%%%%%%%%%%%%%%%%%%%%%%%%%%%%%%%%%%%%%
% \subsection{Heatmaps}

% We illustrate the heatmap from the testing set in the ECU dataset in Figure \ref{fig:heatmap}. Each image in the testing set was converted into HSV color space, and its pixels were allocated into different bins according to the (S, V), (S, H), and (V, H) value tuple. These heatmaps demonstrate the distribution of skin pixels of different skin tones in HSV color space. 

% Take the S-V dimension (Row 1) as an example. From skin tone 1 to skin tone 6, with the skin going darker, the maximum of value channel rises, and the center of the saturation channel moves left. While the heatmap in column seven shows that the training set can not adapt to these changes. After we do color augmentation, the point cloud covers a larger area, so the experimental results are better.

\section{Results on the RFW dataset}
\subsection{Qualitative results}
\begin{figure*}
\begin{center}
   \includegraphics[width=1\linewidth]{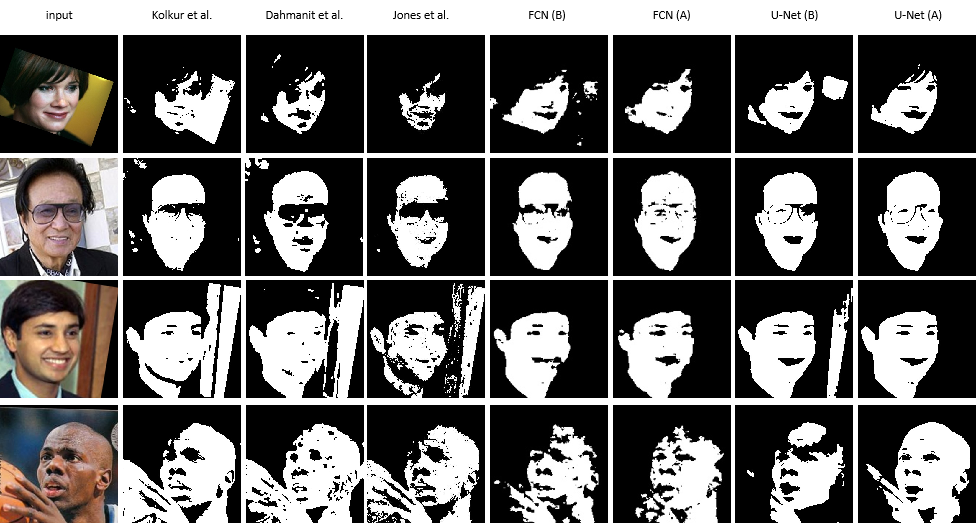}
\end{center}
   \caption{Testing results on the RFW dataset, by various skin segmentation methods including Kolkur ~\etal \cite{kolkur2017human}, Dahmani~\etal\cite{Dahmani}, Jones~\etal\cite{jones2002statistical}, FCN before (B) and after (A) augmentation, and U-Net before (B) and after (A) augmentation (Columns 2 to 8).
Input are shown in column 1.
Result shown for different races: Caucasian, Asian, Indian, and African (Row 1 to 4).}
\label{fig:more results rfw}
\end{figure*}

We demonstrate more results from the RFW dataset in Figure \ref{fig:more results rfw}. The three traditional skin segmentation methods still misclassify the color-like background to be skin pixels. For example, the background of the door is classified as skin areas in the third row by the three baseline methods. On the opposite, in the second row, glasses covered area is not classified as skin areas. What's more, the skin area covered by other items is detected as non-skin pixels.
Compared with results after color augmentation, models without color augmentation are more likely to make false-positive judgments. Moreover, in the darker skin group, models after color augmentation can detect more skin pixels. For example, the result from FCN (A) has less false positive noise than that from FCN (B) in the first row. U-Net (A) detects more skin pixels on the man's head in the fourth row.

\subsection{Skin/face ratios}
\begin{figure*}
\centering
\begin{subfigure}{0.18\textwidth}
\centering
\includegraphics[width=1\linewidth]{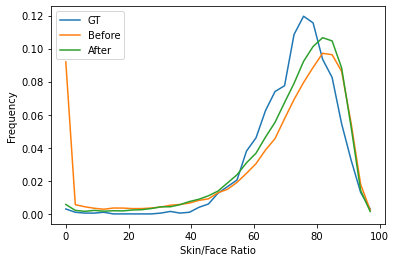}
\caption{Overall}
\end{subfigure}
\begin{subfigure}{0.18\textwidth}
\centering
\includegraphics[width=1\linewidth]{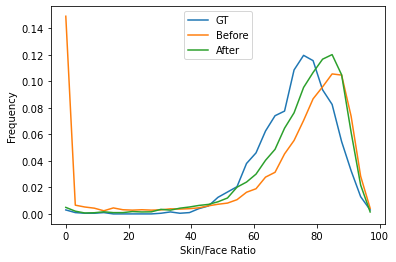}
\caption{Caucasian}
\end{subfigure}
\begin{subfigure}{0.18\textwidth}
\centering
\includegraphics[width=1\linewidth]{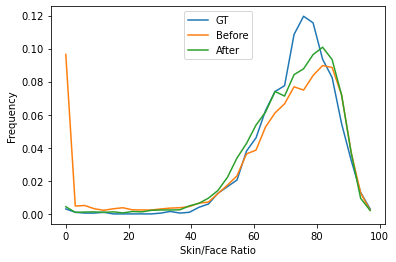}
\caption{Asian}
\end{subfigure}
\begin{subfigure}{0.18\textwidth}
\centering
\includegraphics[width=1\linewidth]{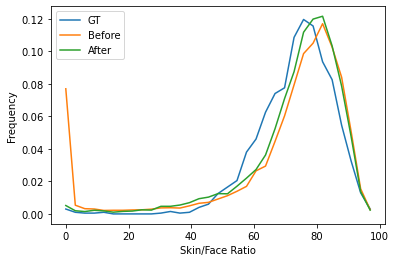}
\caption{Indian}
\end{subfigure}
\begin{subfigure}{0.18\textwidth}
\centering
\includegraphics[width=1\linewidth]{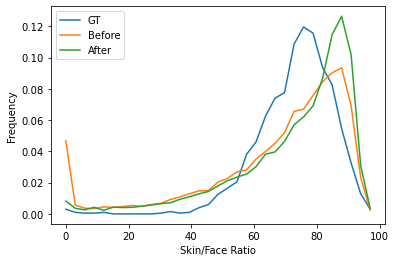}
\caption{African}
\end{subfigure}
\caption{Skin/face ratio distributions curves for Overall RFW dataset (a) and the four different races in RFW dataset (b to e). Blue line refers to the sample distribution curve we get from the annotated ECU dataset. Orange and green line refer to the distribution from testing results before and after color augmentation.}
\label{fig:skin/face ratio distribution}
\end{figure*}

In the main paper, we propose a new method, skin/face ratio, to evaluate the performance of the skin segmentation system with the RFW dataset. It refers to the number of detected skin pixels inside the face area. Although the level of this indicator can reflect the ability of the detection system, larger values do not mean better prediction entirely. In Figure \ref{fig:skin/face ratio distribution}, we extract the skin/face values from various groups of predictions to make it more convincing. 
In this section, we plot skin/face ratio curve to evaluate the performance of the skin segmentation models. The skin/face ratio curve refers to the probability distribution of the skin/face values from the results.

First, we plot skin/face ratio curve using the annotated ECU dataset and its corresponding ground truth, which will be regarded as a sample or a standard (blue). Then, we plot the exact curve of the results from U-Net before and after color augmentation with RFW dataset. The curves are shown in Figure \ref{fig:skin/face ratio distribution}.
We calculate Kullback–Leibler divergence ($D_{KL}$) to measure the difference between the standard probability distribution and that from estimated methods. We expect the resulting curve from a better model to be more relevant to the standard curve, that is, has smaller $D_{KL}$ to the standard distribution.

The $D_{KL}$ values are listed in Table \ref{table:kl div}. It demonstrates that model after color augmentation is more relevant to the standard distribution in Caucasian, Asian, and Indian groups since they have the smaller $D_{KL}$. This also happens in the whole RFW dataset. However, for the African group, the model before color augmentation has a better performance.

From Figure \ref{fig:skin/face ratio distribution}, we find that there is a peak at point '0' for the model before augmentation, which does not appear in the standard curve. This peak indicates that the model does not detect any skin pixels from the face area, which is incorrect. 
After color augmentation, the model works well, and this peak disappears. This explains a lot why skin/face ratio increase after color augmentation. Although the level of this indicator can reflect the ability of the work

\begin{table}
% \small
\begin{center}
% \vspace{-.5cm}
\caption{
Kullback–Leibler divergence between the standard probability distribution and that from estimated methods. Results are from U-Net before and after color augmentation with different groups and the whole RFW dataset.
}
\label{table:kl div}
% \vspace{-.2cm}
\begin{tabular}{c|ccccc}
\hline
 & Cau & Asian & Ind &Afr &Overall \\
\hline
Before Aug.  &0.32 &0.14 &0.15 &0.19  &0.16 \\
After Aug.  &0.05 &0.03 &0.06 &0.22 &0.05	\\
\hline
\end{tabular}
\end{center}
\end{table}
%%%%%%%%%%%%%%%%%%%%%%%%%%%%%%%%%%%%%%%%%%%%%%%%%%%%%%%%%%%%%%%%%%%%%%%%%%%%%%%%%%%%%%%%%%%%%%%%%%%%%%%
%%%%%%%%%%%%%%%%%%%%%%%%%%%%%%%%%%%%%%%%%%%%%%%%%%%%%%%%%%%%%%%%%%%%%%%%%%%%%%%%%%%%%%%%%%%%%%%%%%%%%%%
\section{Results from cross dataset experiments}
In this section we illustrate additional results of grayscale images from the ECU dataset and our self-made drastic dataset.
FCN (B) can detect only a small area of skin pixel in grayscale image but U-Net (B) fail to detect any skin pixels.
On the other hand, both models without color augmentation can hardly detect a single skin pixel for the images with unconstrained illumination. 
Improvements are obvious after color augmentation is applied to the network.
Models with color augmentation work well and correctly detected skin pixels for both grayscale images and images with unconstrained illumination.
%%%%%%%%%%%%%%%%%%%%%%%%%%%%%%%%%%%%%%%%%%%%%%%%%%%%%%%%%%%%%%%%%%%%%%%%%%%%%%%%%%%%%%%%%%%%%%%%%%%%%%%
%%%%%%%%%%%%%%%%%%%%%%%%%%%%%%%%%%%%%%%%%%%%%%%%%%%%%%%%%%%%%%%%%%%%%%%%%%%%%%%%%%%%%%%%%%%%%%%%%%%%%%%
%%%%%%%%%%%%%%%%%%%%%%%%%%%%%%%%%%%%%%%%%%%%%%%%%%%%%%%%%%%%%%%%%%%%%%%%%%%%%%%%%%%%%%%%%%%%%%%%%%%%%%%
%%%%%%%%%%%%%%%%%%%%%%%%%%%%%%%%%%%%%%%%%%%%%%%%%%%%%%%%%%%%%%%%%%%%%%%%%%%%%%%%%%%%%%%%%%%%%%%%%%%%%%%
%%%%%%%%%%%%%%%%%%%%%%%%%%%%%%%%%%%%%%%%%%%%%%%%%%%%%%%%%%%%%%%%%%%%%%%%%%%%%%%%%%%%%%%%%%%%%%%%%%%%%%%
% \begin{figure*}
% \begin{subfigure}{1\textwidth}
% \includegraphics[width=1\linewidth]{figures/heatmap ecu sv.png}
% \caption{}
% \end{subfigure}
% \begin{subfigure}{1\textwidth}
% \includegraphics[width=1\linewidth]{figures/heatmap ecu sh.png}
% \caption{}
% \end{subfigure}
% \begin{subfigure}{1\textwidth}
% \includegraphics[width=1\linewidth]{figures/heatmap ecu vh.png}
% \caption{}
% \end{subfigure}
% \caption{Heatmaps from ECU dataset in three dimensions: (a)Saturation-Value, (b)Saturation-Hue, (c) Value-Hue. 
% The first six columns mark the skin pixels distributions of Fitzpatrick\cite{fitzpatrick1988validity} skin tone I-VI. The last two columns refer to the skin pixels distribution of the training set before and after our color space augmentation.}
% \label{fig:heatmap}
% \end{figure*}
%%%%%%%%%%%%%%%%%%%%%%%%%%%%%%%%%%%%%%%%%%%%%%%%%%%%%%%%%%%%%%%%%%%%%%%%%%%%%%%%%%%%%%%%%%%%%%%%%%%%%%%%%%%%%%%%%%%%%%%%%%%%%%%%%%%%%%%%%%%%%%%%%%%%%%%%%%%%%%%%%%%%%%%%%%%%%%%%%%%%%%%%%%%%%%%%%%%%%%%%%%%%%%
\begin{figure*}
% \centering
\begin{subfigure}{0.49\textwidth}
\centering
\includegraphics[width=0.88\linewidth]{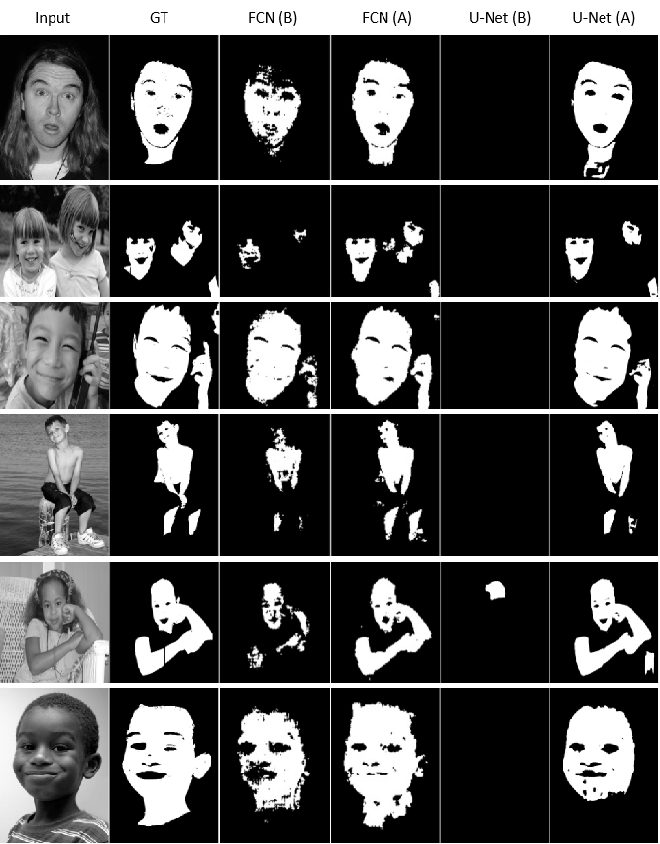}
\caption{}
\end{subfigure}
\begin{subfigure}{0.49\textwidth}
\centering
\includegraphics[width=1\linewidth]{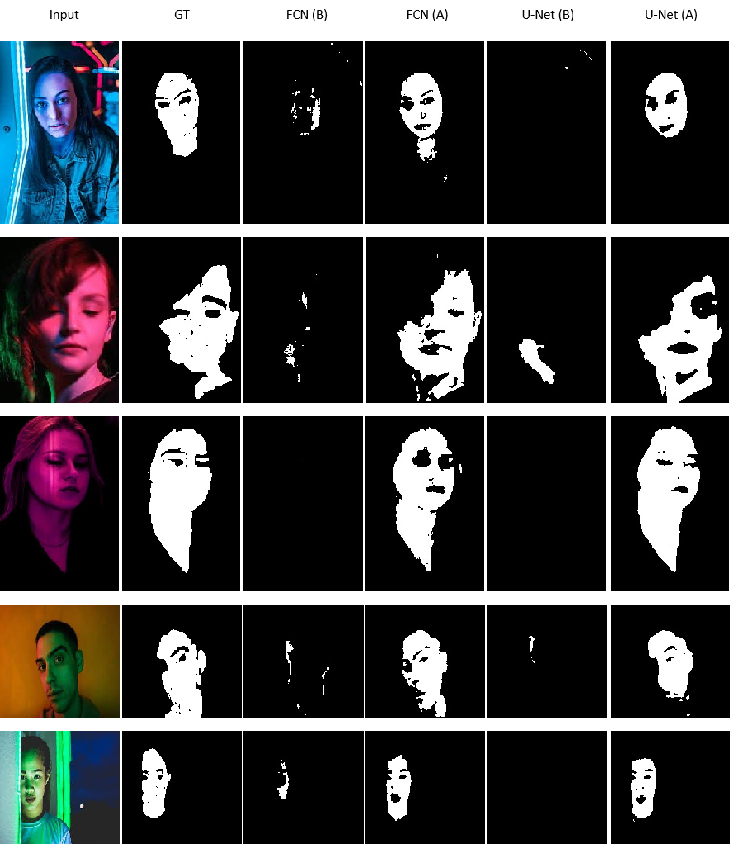}
\caption{}
\end{subfigure}

\caption{Testing results on our self-made dataset (a) and grayscale images from the ECU dataset (B) by deep learning models FCN and U-Net. The label (B) on the top of the images refers to the results from the model before color augmentation. In comparison, label (A) refers to the model with color augmentation. 
Input images and ground truth are shown in columns 1 and 2 in each group.
}
\label{fig:more light gray results}
\end{figure*}
%%%%%%%%%%%%%%%%%%%%%%%%%%%%%%%%%%%%%%%%%%%%%%%%%%%%%%%%%%%%%%%%%%%%%%%%%%%%%%%%%%%%%%%%%%%%%%%%%%%%%%

%%%%%%%%% REFERENCES
{\small
\bibliographystyle{ieee_fullname}
\bibliography{supp}
}